\documentclass[10pt,twocolumn,letterpaper]{article}

\usepackage[pagenumbers]{cvpr} 

\definecolor{cvprblue}{rgb}{0.21,0.49,0.74}
\usepackage[accsupp]{axessibility}
\usepackage[pagebackref,breaklinks,colorlinks,allcolors=cvprblue]{hyperref}
\usepackage[utf8]{inputenc}
\usepackage[T1]{fontenc}
\usepackage[inline]{enumitem}
\usepackage{amsfonts}       
\usepackage{amsmath}
\usepackage{bm}
\usepackage{booktabs}       
\usepackage{tabularx}
\usepackage{times}
\usepackage{epsfig}
\usepackage{graphicx}
\usepackage{multirow}
\usepackage{makecell}
\usepackage{nicefrac}       
\usepackage{xcolor}         
\usepackage{algorithm}
\usepackage{algorithmic}
\usepackage{url}
\usepackage{etoolbox,lipsum}
\usepackage{indentfirst}
\usepackage[dvipsnames]{xcolor}
\usepackage{xspace}
\usepackage{bbm}
\usepackage{amssymb}
\usepackage{microtype}
\usepackage{pifont}
\usepackage{cuted}
\def\paperID{7744} 
\def\confName{CVPR}
\def\confYear{2025}
\newcommand\blfootnote[1]{%
  \begingroup
  \renewcommand\thefootnote{}\footnote{#1}%
  \addtocounter{footnote}{-1}%
  \endgroup
}
\newcommand{\ourmethod}{\textup{DeSiRe-GS}\xspace}

\title{\ourmethod: 4D Street Gaussians for Static-Dynamic \underline{De}composition \\ and \underline{S}urface \underline{Re}construction for Urban Driving Scenes}

\author{Chensheng Peng$^*$ \quad
Chengwei Zhang$^*$ \quad Yixiao Wang\quad Chenfeng Xu\quad Yichen Xie\\ Wenzhao Zheng\quad Kurt Keutzer\quad Masayoshi~Tomizuka\quad Wei Zhan 
\vspace{0.2cm} \\
{UC Berkeley}
}

\begin{document}

\twocolumn[{%
\renewcommand\twocolumn[1][]{#1}%
\maketitle
\vspace{-0.5cm}
\begin{center}
    \centering
    \captionsetup{type=figure}
    \includegraphics[width=\textwidth]{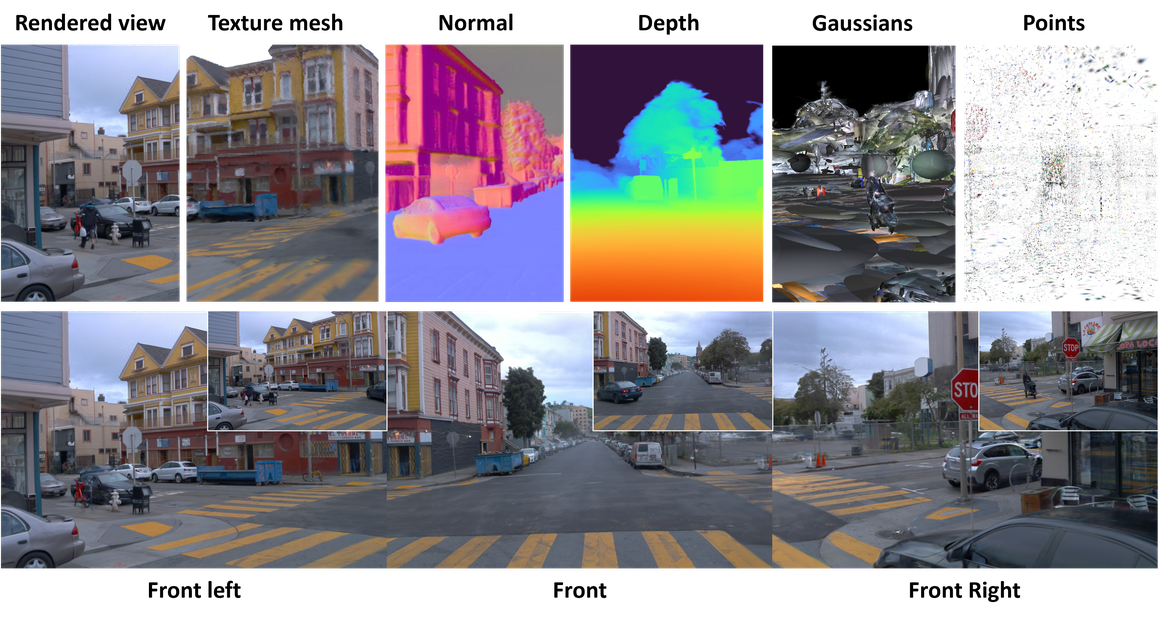}
    \captionof{figure}{\textbf{\ourmethod}. We present a 4D street gaussian splatting representation for self-supervised static-dynamic decomposition and high-fidelity surface reconstruction without the requirement for extra 3D annotations such as bounding boxes. }
    \label{fig:teaser}
\end{center}%
}]

\blfootnote{$^*$Equal Contribution.}
\begin{abstract}

We present \ourmethod, a self-supervised gaussian splatting representation, enabling effective static-dynamic decomposition and high-fidelity surface reconstruction in complex driving scenarios. Our approach employs a two-stage optimization pipeline of dynamic street Gaussians. In the first stage, we extract 2D motion masks based on the observation that 3D Gaussian Splatting inherently can  reconstruct only the static regions in dynamic environments. These extracted 2D motion priors are then mapped into the Gaussian space in a differentiable manner, leveraging an efficient formulation of dynamic Gaussians in the second stage. Combined with the introduced geometric regularizations, our method are able to address the over-fitting issues caused by data sparsity in autonomous driving, reconstructing physically plausible Gaussians that align with object surfaces  rather than floating in air. Furthermore, we introduce  temporal cross-view consistency to ensure coherence across time and viewpoints, resulting in high-quality surface reconstruction. Comprehensive experiments demonstrate the efficiency and effectiveness of \ourmethod, surpassing prior self-supervised arts and achieving accuracy comparable to methods relying on external 3D bounding box annotations. Code is available 
at 
\url{https://github.com/chengweialan/DeSiRe-GS}
\end{abstract}


\section{Introduction}
\label{sec:intro}

Modeling driving scenes \cite{kitti_2012_cvpr, waymo_2020_cvpr} is essential for autonomous driving applications \cite{pnasmot, wang2023interactive}, as it facilitates real-world simulation and supports scene understanding \cite{hugs_2024_cvpr}. An effective scene representation enables a system to efficiently perceive and reconstruct dynamic driving environments.
Recent 3D Gaussian Splatting (3DGS) \cite{3dgs_2023_TOG} has emerged as a prominent 3D representation that can be optimized through 2D supervision. It has gained popularity due to its explicit nature, high efficiency, and rendering speed. 

While 3D Gaussian Splatting (3DGS) has demonstrated strong performance in static object-centric reconstructions, the original 3DGS struggles to handle dynamic objects in unbounded street views, which are common in real-world scenarios, particularly for autonomous driving applications. It is unable to effectively model dynamic regions, leading to blurring artifacts due to the Gaussian model's time-independent parameterization. As a result, 4D-GS~\cite{4dgs_2024_cvpr} is proposed, modeling the dynamics with a Hexplane encoder. The Hexplane \cite{hexplane_2023_cvpr} works well on object-level datasets, but struggles with driving scenes because of the unbounded areas in outdoor environments. Instead, we choose to reformulate the original static Gaussian model as time-dependent variables with minor changes, ensuring the efficiency of handling large-scale driving scenes.



In this paper, we present \ourmethod, a purely Gaussian Splatting-based representation, which facilitates self-supervised static-dynamic decomposition and high-quality surface reconstruction in driving scenarios.
For static-dynamic decomposition, existing methods such as DrivingGaussian \cite{drivinggaussian_2024_cvpr} and Street Gaussians \cite{streetgs_2024_eccv}, rely on explicit 3D bounding boxes, which significantly simplifies the decomposition problem, since dynamic Gaussians in a moving bounding box can be simply removed. Without the 3D annotations, some recent self-supervised methods like PVG \cite{pvg_2023_arxiv} and S3Gaussian \cite{s3g_2024_arxiv} have attempted to achieve decomposition but fall short in performance, as they treat all Gaussians as dynamic, relying on indirect supervision to learn motion patterns. However, our proposed method can achieve effective self-supervised decomposition, based on a simple observation that dynamic regions reconstructed from 3DGS are blurry---quite different from the ground truth images. Despite the absence of 3D annotations, \ourmethod produces results comparable to, or better than, approaches that use explicit bounding boxes for decomposition.

Another challenge in applying 3D Gaussian Splatting (3DGS) to autonomous driving is the sparse nature of images, which is more pronounced compared to object-centric reconstruction tasks. This sparsity often leads 3DGS to overfit on the limited number of observations, resulting in inaccurate geometry learning. Inspired by 2D Gaussian Splatting (2DGS) \cite{2dgs_2024_siggraph}, we aim to generate flatter, disk-shaped Gaussians to better align with the surfaces of objects like roads and walls. We also couple the normal and scale of each Gaussian, which can be optimized jointly to improve surface reconstruction quality. 

To further address the overfitting issue, we propose temporal geometrical cross-view consistency, which significantly enhances the model’s geometric awareness and accuracy by aggregating information from different views across time. These strategies allow us to achieve state-of-the-art reconstruction quality, surpassing other Gaussian splatting approaches in the field of autonomous driving.

Overall, \ourmethod makes the following contributions:

\begin{itemize}
    \item We propose to extract motion information easily from appearance differences based on a simple observation that 3DGS cannot successfully model the dynamic regions.
    \item We then distill the extracted 2D motion priors in local frames into global gaussian space, using time-varying Gaussians in a differentiable manner.
    \item We introduce effective 3D regularizations and temporal cross-view consistency to generate physically reasonable Gaussian ellipsoids, further enhancing high-quality decomposition and reconstruction.
\end{itemize}

We demonstrate \ourmethod's capability of effective static-dynamic decomposition and high-fidelity surface reconstruction across various challenging datasets \cite{waymo_2020_cvpr, kitti_2012_cvpr}.

\begin{figure*}[th!]
    \centering
    \includegraphics[width=\linewidth]{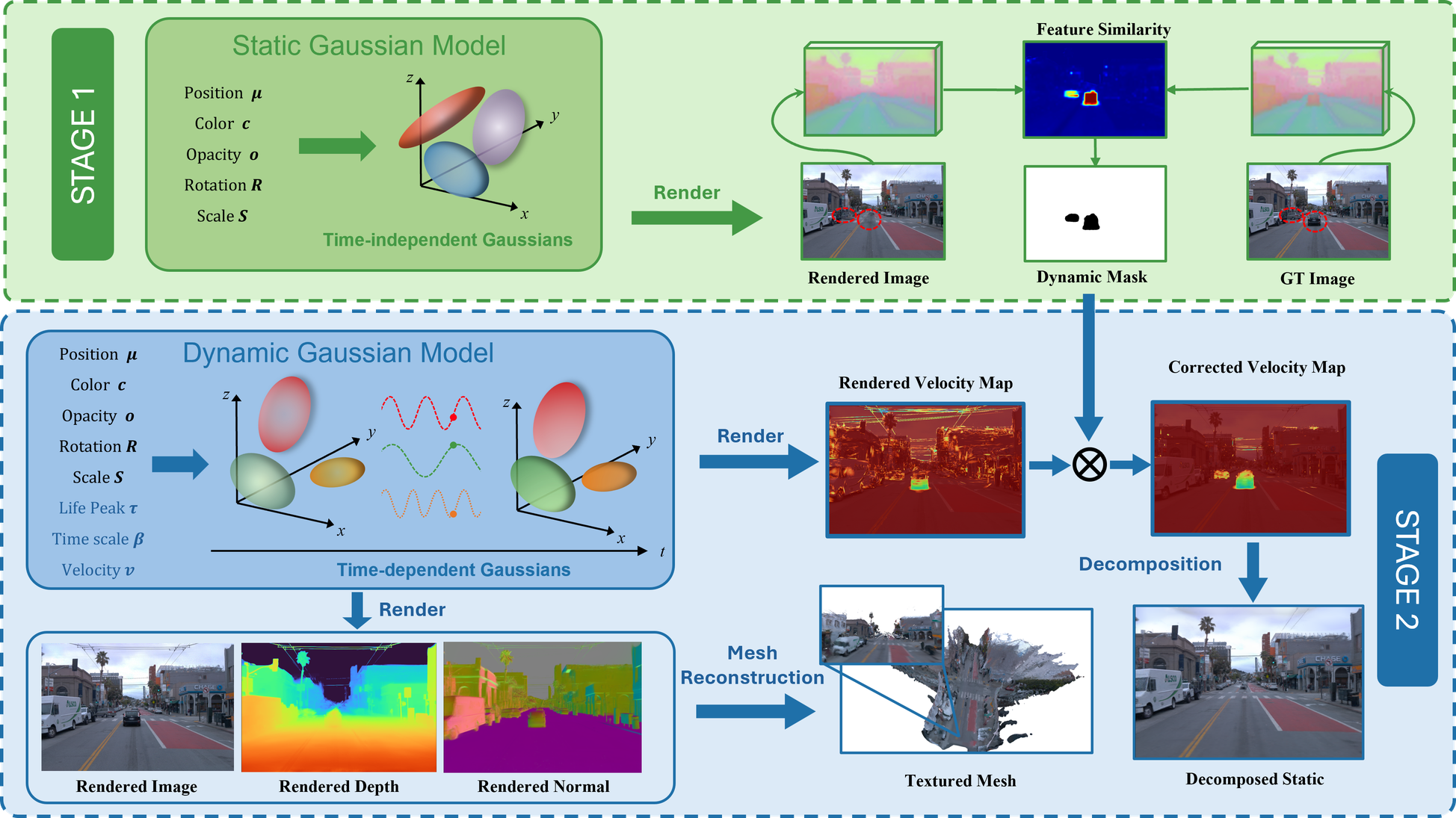}
    \caption{Pipeline of \ourmethod. To tackle the challenges in self-supervised street scene decomposition. The
entire pipeline is optimized without extra annotations in a self-supervised manner, leading to superior
scene decomposition ability and rendering quality.}
    \label{fig:pipeline}
    \vspace{-2pt}
\end{figure*}

\section{Related Work}
\label{sec:related_work}

\textbf{Urban Scene Reconstruction.} Recent advancements in 3D representation, such as point cloud \cite{peng2023delflow, liu2024dvlo, liu2024point}, Neural Radiance Field (NeRF) \cite{nerf_2020_eccv} and 3D Gaussian Splatting (3DGS) \cite{3dgs_2023_TOG}, have significantly advanced urban scene reconstruction. Many studies \cite{urbannerf_2022_cvpr,blocknerf_2022_cvpr,meganerf_2022_cvpr,emernerf_2023_arxiv,rodus_2024_eccv,nsg_2021_cvpr,mars_2023_caai, qslam} have integrated NeRF into workflows for autonomous driving. Urban Radiance Fields \cite{urbannerf_2022_cvpr} combine lidar and RGB data, while Block-NeRF \cite{blocknerf_2022_cvpr} and Mega-NeRF \cite{meganerf_2022_cvpr} partition large scenes for parallel training. However, dynamic environments pose challenges. NSG \cite{nsg_2021_cvpr}, use neural scene graphs to decompose dynamic scenes, and SUDS \cite{suds_2023_cvpr} introduces a multi-branch hash table for 4D scene representation. Self-supervised approaches like EmerNeRF \cite{emernerf_2023_arxiv} and RoDUS \cite{rodus_2024_eccv} can effectively address dynamic scene challenges. EmerNeRF capturing object correspondences via scene flow estimation, and RoDUS utilizes a robust kernel-based training strategy combined with semantic supervision.

In 3DGS-based urban reconstruction, recent works \cite{streetgs_2024_eccv,drivinggaussian_2024_cvpr,omnire_2024_arxiv,hugs_2024_cvpr,pvg_2023_arxiv,s3g_2024_arxiv} have gained attention. StreetGaussians \cite{streetgs_2024_eccv} models static and dynamic scenes separately using spherical harmonics, while DrivingGaussian \cite{drivinggaussian_2024_cvpr} introduces specific modules for static background and dynamic object reconstruction. OmniRe \cite{omnire_2024_arxiv} unifies static and dynamic object reconstruction via dynamic Gaussian scene graphs. However, \cite{streetgs_2024_eccv, omnire_2024_arxiv, hugs_2024_cvpr} all require additional 3D bounding boxes which are sometimes difficult to obtain.



\textbf{Static Dynamic Decomposition.} Several approaches seek to model the deformation of dynamic and static components. D-NeRF \cite{dnerf_cvpr_2020}, Nerfiles \cite{nerfiles_iccv_2021}, Deformable GS \cite{deformgs_2024_cvpr} and 4D-GS \cite{4dgs_2024_cvpr} extend the vanilla NeRF or 3DGS by incorporating a deformation field. They compute the canonical-to-observation transformation and separate static and dynamic components through the deformation network. However, applying such methods to large-scale driving scenarios is challenging due to the substantial computational resources needed for learning dense deformation parameters, and the inaccurate decomposition leads to suboptimal peformance. 


For autonomous driving scenarios, NSG \cite{nsg_2021_cvpr} models dynamic and static parts as nodes in neural scene graphs but requires additional 3D annotations. Other NeRF-based methods \cite{suds_2023_cvpr,rodus_2024_eccv,emernerf_2023_arxiv} leverage a multi-branch structure to train time-dependent and time-invariant features separately. 3DGS-based methods, such as \cite{pvg_2023_arxiv,streetgs_2024_eccv,drivinggaussian_2024_cvpr,s3g_2024_arxiv}, also focus on static-dynamic separation but still face limitations. \cite{s3g_2024_arxiv} utilizes a deformation network with a hexplane temporal-spatial encoder, requiring extensive computation. PVG \cite{pvg_2023_arxiv} assigns attributes like velocity and lifespan to each Gaussian, distinguishing static from dynamic ones. Yet, the separation remains incomplete and lacks thoroughness.

\textbf{Neural Surface Reconstruction.} 
Traditional methods for Neural Surface Reconstruction focus more on real geometry structures. With the rise of neural radiance field (NeRF) technologies, neural implicit representations have shown promise for high-fidelity surface reconstruction. Approaches like \cite{neuralangelo_2023_cvpr, monosdf_2022_cvpr, volsdf_2021_cvpr, neus_2021_arxiv} train neural signed distance functions (SDF) to represent scenes. StreetSurf \cite{streetsurf_2023_arxiv} proposes disentangling close and distant views for better implicit surface reconstruction in urban settings, while \cite{urbannerf_2022_cvpr} steps further using sparse lidar to enhance depth details.

3D GS has renewed interest in explicit geometric reconstruction, with recent works \cite{dnsplatter_2024_cvpr, sugar_2024_cvpr, 2dgs_2024_siggraph, neusg_2023_cvpr, trim3d_2024_cvpr, gsdf_2024_cvpr, pgsr_2024_arxiv} focusing on geometric regularization techniques. SuGaR \cite{sugar_2024_cvpr} aligns Gaussian ellipsoids to object surfaces through introducing and additional regularization term, while 2DGS \cite{2dgs_2024_siggraph} directly replaces 3D ellipsoids with 2D discs and utilizes the truncated signed distance function (TSDF) to fuse depth maps, enabling noise-free surface reconstruction. PGSR \cite{pgsr_2024_arxiv} introduces single- and multi-view regularization for multi-view consistency. GSDF \cite{gsdf_2024_cvpr} and NeuSG \cite{neusg_2023_cvpr} combine 3D Gaussians with neural implicit SDFs to enhance surface details. TrimGS \cite{trim3d_2024_cvpr} refines surface structures by trimming inaccurate geometry, maintaining compatibility with earlier methods like 3DGS and 2DGS. While these approaches excel in small-scale reconstruction, newer works like \cite{gigags_2024_arxiv, gaussianpro_2024_icml, rogs_2024_cvpr} aim to address large-scale urban scenes. \cite{gigags_2024_arxiv} adopts a large-scene partitioning strategy for reconstruction, while RoGS \cite{rogs_2024_cvpr} proposes 2D Gaussian surfel representation which aligns with physical characteristics of road surfaces.




\section{Preliminary}
\label{sec:prelim}
\noindent \textbf{3D Gaussian Splatting:} 3D Gaussian Splatting (3DGS) \cite{3dgs_2023_TOG} employs a collection of colored ellipsoids, $G = \{g\}$ to explicitly represent 3D scenes. Each Gaussian \(g = \{\boldsymbol{\mu}, \boldsymbol{s}, \boldsymbol{r}, o, \boldsymbol{c}\}\) is defined by the following learnable attributes: a position center \( \boldsymbol{\mu} \in \mathbb{R}^3 \), a covariance matrix \( \boldsymbol{\Sigma} \in \mathbb{R}^{3 \times 3} \), an opacity scalar \( o \), and a color vector \( \boldsymbol{c} \), which is modeled using spherical harmonics. The distribution of 3D Gaussians is mathematically described as:
\begin{equation}
    G(\boldsymbol{x}) = \exp \left\{ -\frac{1}{2} (\boldsymbol{x} - \boldsymbol{\mu})^\top \boldsymbol{\Sigma}^{-1} (\boldsymbol{x} - \boldsymbol{\mu}) \right\}.
\end{equation}
The covariance matrix \(\boldsymbol{\Sigma}\) can be formulated as follows:
$\boldsymbol{\Sigma} = \mathbf{R} \mathbf{S} \mathbf{S}^{\top} \mathbf{R}^{\top}$, where
\(\mathbf{S} \) denotes a diagonal scaling matrix, while \( \mathbf{R} \) is a rotation matrix, parameterized as a scaling vector \( \boldsymbol{s} \) and a quaternion \( \boldsymbol{r} \in \mathbb{R}^4 \), respectively.

To generate images from a specific viewpoint, 3D gaussian ellipsoids are projected onto a 2D image plane to form 2D ellipses for rendering. 
For each pixel, a sequence of Gaussians \( \mathcal{N} \) is sorted in ascending order based on depth. The color is then rendered through alpha blending:
\begin{equation}
    C = \sum_{i \in \mathcal{N}} c_i \alpha_i \prod_{j=1}^{i-1} (1 - \alpha_j),
\end{equation}
where \(\alpha_i\) and \(c_i\) denote the density and color of the \(i\)-\textit{th} Gaussian, respectively, derived from the learned opacity and SH coefficients of the corresponding Gaussian.

\vspace{5pt}
\noindent \textbf{Periodic Vibration Gaussian (PVG): } PVG \cite{pvg_2023_arxiv} reshapes the original Gaussian model by introducing time-dependent adjustments to the position mean \(\boldsymbol{\mu}\) and opacity \(o\). The modified model is represented as follows:
\begin{equation}
    \tilde{\boldsymbol{\mu}}(t) = \boldsymbol{\mu} + \frac{l}{2\pi} \cdot \sin\left( \frac{2\pi (t - \tau)}{l} \right) \cdot \boldsymbol{v},
\end{equation}
\begin{equation}
    \tilde{o}(t) = o \cdot e^{-\frac{1}{2} (t - \tau)^2 \beta^{-2}},
\end{equation}
where \(\tilde{\boldsymbol{\mu}}(t)\) denotes vibrating position centered at \(\boldsymbol{\mu}\) occurring around life peak \(\tau\), and \(\tilde{o}(t)\) represents the time-dependent opacity which undergoes exponential decay as time deviates from the life peak \(\tau\). \(\beta\) and \(\boldsymbol{v}\) determine the decay rate and the instant velocity at the life peak \(\tau\) , respectively, and are both learnable parameters. \(l\), as a pre-defined parameter of the scene, represents the oscillation period. Thus, the PVG model is expressed as:
\begin{equation}
    \mathcal{G}(t) = \{\tilde{\boldsymbol{\mu}}(t), \boldsymbol{s}, \boldsymbol{r},  \tilde{o}(t), \boldsymbol{c}, \tau, \beta, \boldsymbol{v}\},
\end{equation}
 
We adopt PVG as the dynamic representation for  autonomous driving scenes, because PVG model preserves the structure of the original 3D GS model at any given time $t$,  enabling it to be rendered using the standard 3D GS pipeline to reconstruct the dynamic scene.  For further details about PVG, we refer the readers to \cite{pvg_2023_arxiv}.

\section{\ourmethod}

As shown in Fig. \ref{fig:pipeline}, the training process is divided into two stages. We first extract 2D motion masks by calculating the feature difference between the rendered image and GT image. In the second stage, we distill the 2D motion information into Gaussian space using PVG \cite{pvg_2023_arxiv}, enabling the rectification of inaccurate attributes for each Gaussian in a differentiable manner. 

\subsection{Dynamic Mask Extraction (stage I)}
During the first stage, we observe that 3D Gaussian Splatting (3DGS) performs effectively in reconstructing static elements, such as parked cars and buildings in a driving scene. However, it struggles to accurately reconstruct dynamic regions, as the original 3DGS does not incorporate temporal information. This limitation results in artifacts such as ghost-like floating points in the rendered images, as illustrated in Fig. \ref{fig:pipeline} (stage 1). To address this issue, we leverage the significant differences between static and dynamic regions to develop an efficient method for extracting segmentation masks that encode motion information. 

Initially, a pretrained foundation model is employed to extract features from both the rendered image and the ground truth (GT) image used for supervision. Let $\hat{F}$ denote the features extracted from the rendered image $\hat{I}$, and $F$ represent the features extracted from the GT image $I$. To distinguish dynamic and static regions, we compute the per-pixel dissimilarity $D$ between the corresponding features. The dissimilarity metric $D$ approaches 0 for similar features, indicating static regions, and nears 1 for dissimilar features, corresponding to dynamic regions.
\begin{equation}
    D = \left( 1 -  \texttt{cos}( \hat{F}, F) \right) / 2.
\end{equation}

As the pretrained model is frozen, the resulting dissimilarity score $D \in R^{H \times W}$ is computed without involving any learnable parameters. Rather than applying a simple threshold to $D$ to generate a motion segmentation mask, we propose a multi-layer perceptron (MLP) decoder to predict the dynamicness $\delta \in R^{H \times W}$. This decoder leverages the extracted features, which contain rich semantic information, while the dissimilarity score is employed to guide and optimize the learning process of the decoder.
\begin{equation}
\label{eq:L-d}
    \mathcal{L}_{dyn} = \delta \odot D,
\end{equation}
where $\odot$ refers to the element-wise multiplication.

By employing the loss function $\mathcal{L}_{dyn}$ defined in Eq. \ref{eq:L-d}, the decoder is optimized to predict lower values in regions where $D$ is high, corresponding to dynamic regions, thereby minimizing the loss. We can then obtain the binary mask encoding motion information ($\varepsilon $ is a fixed threshold):
\begin{equation}
    M = \mathbb{I} \left( \delta > \varepsilon  \right).
\end{equation}

During training, the joint optimization of image rendering and mask prediction mutually benefits each other. By excluding dynamic regions during supervision, the differences between rendered images and GT images become more noticeable, facilitating the extraction of motion masks.
\begin{equation}
    \mathcal{L}_{masked-render} = M \odot\| \hat{I} - I \|.
\end{equation}

\subsection{Static Dynamic Decomposition (stage II)}
While stage I provides effective dynamic masks, these masks are confined to the image space rather than the 3D Gaussian space and depend on ground truth images. This reliance limits their applicability in novel view synthesis, where supervised images may not be available.

To bridge the 2D motion information from stage I to the 3D Gaussian space, we adopt PVG, a unified representation for dynamic scenes (Section \ref{sec:prelim}). However, PVG's reliance on image and sparse depth map supervision introduces challenges, as accurate motion patterns are difficult to learn from indirect supervision signals. Consequently, the rendered velocity map $\mathbf{V} \in R^{H \times W}$, as shown in Fig. \ref{fig:pipeline} (stage 2), often contains noisy outliers. For example, static regions such as roads and buildings, where the velocity should be zero, are not handled effectively. This results in unsatisfactory scene decomposition, with PVG frequently misclassifying regions where zero velocity is expected.

To mitigate this issue and generate more accurate Gaussian representations, we incorporate the segmentation masks obtained from stage I to regularize the 2D velocity map $\mathbf{V}$, which is rendered from Gaussians in the 3D space.
\begin{equation}
\label{eq:vreg}
    \mathcal{L}_v = \mathbf{V} \odot M.
\end{equation}

Minimizing $\mathcal{L}_v$ penalizes regions where velocity should be zero, effectively eliminating noisy outliers produced by the original PVG. This process propagates motion information from the 2D local frame to the global Gaussian space. With the refined velocity 
$v$ for each Gaussian, dynamic and static Gaussians can be distinguished by applying a simple threshold. This approach achieves superior self-supervised decomposition compared to PVG \cite{pvg_2023_arxiv} and S3Gaussian \cite{s3g_2024_arxiv}, without requiring additional 3D annotations such as bounding boxes used in previous methods \cite{hugs_2024_cvpr, streetgs_2024_eccv, omnire_2024_arxiv}.

\subsection{Surface Reconstruction}

\subsubsection{Geometric Regularization}
\noindent{}{\bf Flattening 3D Gaussian:} 
Inspired by 2D Gaussian Splatting (2DGS) \cite{2dgs_2024_siggraph}, we aim to flatten 3D ellipsoids into 2D disks, allowing the optimized Gaussians to better conform to object surfaces and enabling high-quality surface reconstruction. The scale $s = (s_1, s_2, s_3)$ of 3DGS defines the ellipsoid's size along three orthogonal axes. Minimizing the scale along the shortest axis effectively transforms 3D ellipsoids into 2D disks. The scaling regularization loss is:
\begin{equation}
\label{eq:min-scale-reg}
    \mathcal{L}_s = \| \min (s_1, s_2, s_3) \|.
\end{equation}

\begin{figure}
    \centering
\includegraphics[width=\linewidth]{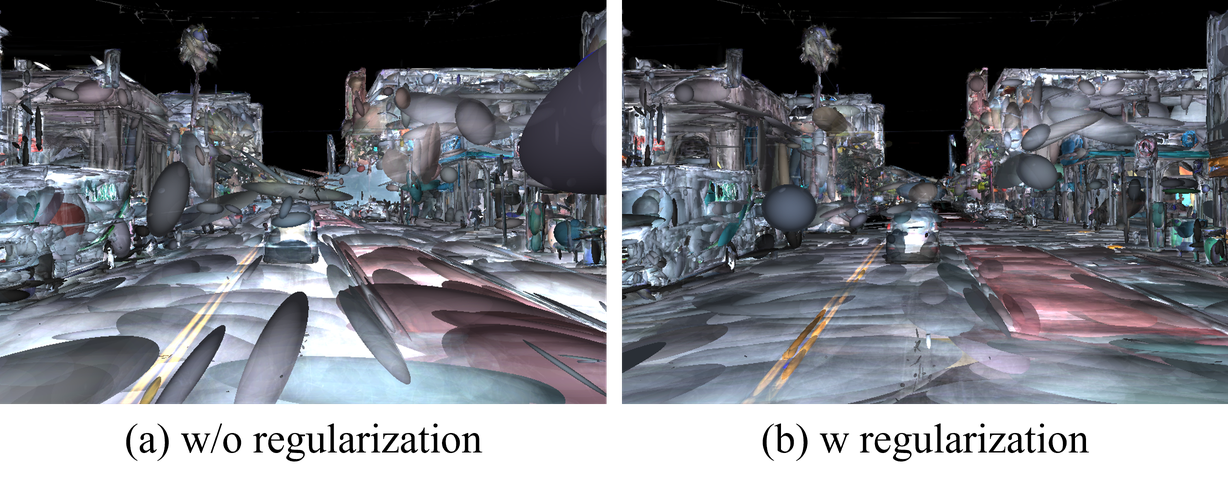}
\vspace{-20pt}
    \caption{Gaussian Scale Regularization.}
    \label{fig:gaint_reg}
\end{figure}

\noindent{}{\bf Normal Derivation:} Surface normals are critical for surface reconstruction. Previous methods incorporate normals by appending a normal vector $n_i \in R^3$ to each Gaussian, which is then used to render a normal map $\mathcal{N} \in R^{H\times W}$.  The ground truth normal map is employed to supervise the optimization of the Gaussian normals. However, these approaches often fail to achieve accurate surface reconstruction, as they overlook the inherent relationship between the scale and the normal. Instead of appending a separate normal vector, we derive the normal $n$ directly from the scale vector $s$. The normal direction naturally aligns with the axis corresponding to the smallest scale component, as the Gaussians are shaped like a disk after flattening regularization.
\begin{equation}
    n = R \cdot \arg \min (s_1, s_2, s_3).
\end{equation}

With such formulation of the normal, the gradient can be back-propagated to the scale vector, rather than the appended normal vector, thereby facilitating better optimization of the Gaussian parameters. The normal loss is:
\begin{equation}
\label{eq:normal}
    \mathcal{L}_n = \| \mathcal{N} - \hat{\mathcal{N}} \|_2.
\end{equation}

\noindent{}{\bf Giant Gaussian Regularization:} We observed that both 3DGS and PVG could produce oversized Gaussian ellipsoids without additional regularization, particularly in unbounded driving scenarios, as illustrated in Fig. \ref{fig:gaint_reg} (a).

 Our primary objective is to fit appropriately scaled Gaussians that support accurate image rendering and surface reconstruction. While oversized Gaussian ellipsoids with low opacity may have minimal impact on the rendered image, they can significantly impair surface reconstruction. This is a limitation often overlooked in existing methods focused solely on 2D image rendering. To address this issue, we introduce a penalty term for each Gaussian:
\begin{equation}
\label{eq:big-scale-reg}
s_g = \max (s_1, s_2, s_3); \quad    \mathcal{L}_g = s_g \cdot  \mathbb{I}(s_g  > \epsilon),
 \end{equation}
where $s_g$ is the largest scale direction and $\epsilon$ is a predefined threshold for huge gaussians.

\begin{figure}
    \centering
    \includegraphics[width=\linewidth]{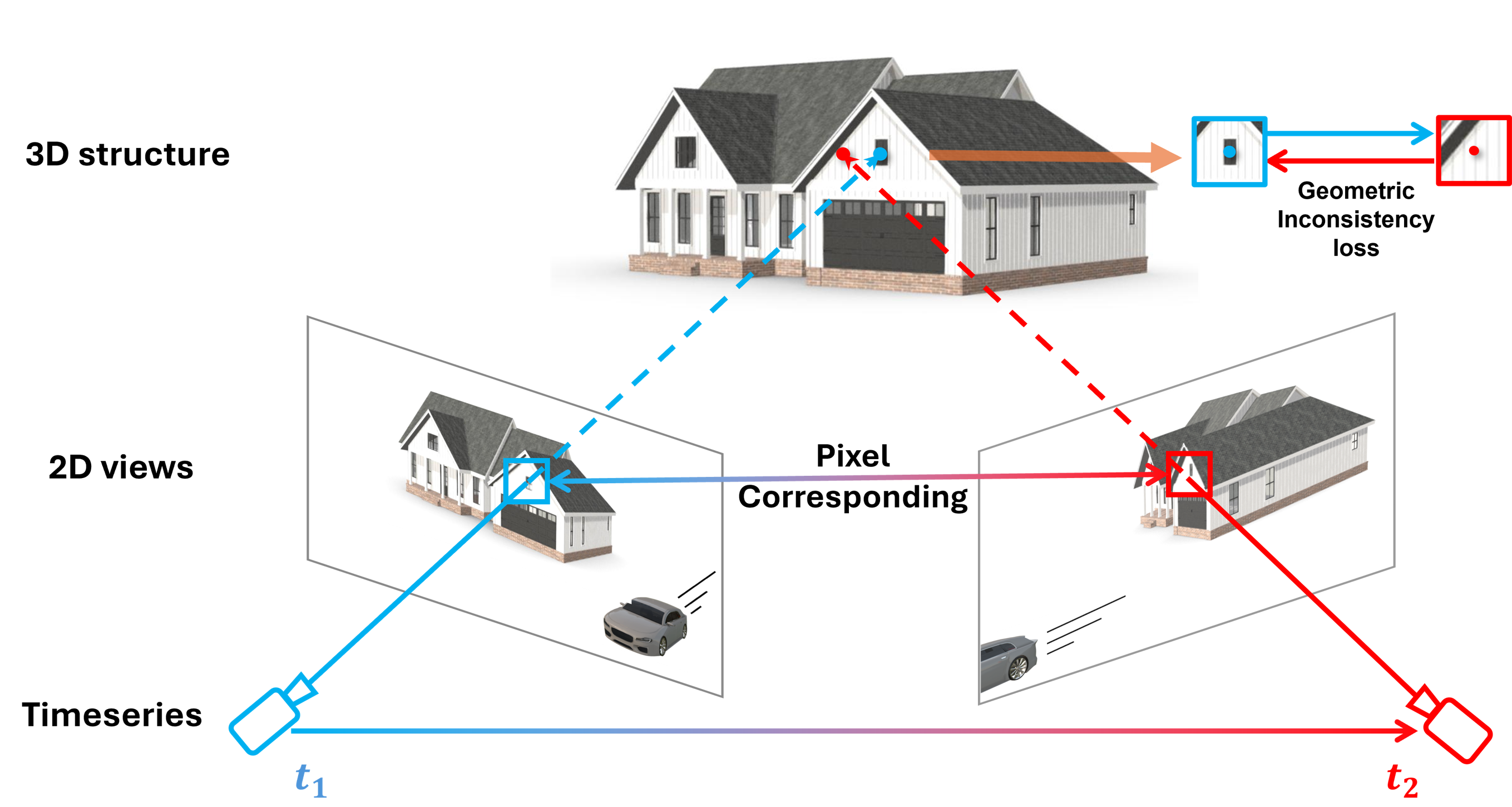}
    \caption{Cross-view consistency}
    \label{fig:multi_view_consistency}
\end{figure}

\subsubsection{Temporal Spatial Consistency}
In driving scenarios, the sparse nature of views often leads to overfitting to the training views during the optimization of Gaussians. Single-view image loss is particularly susceptible to challenges in texture-less regions at far distances. As a result, relying on photometric supervision from images and sparse depth map is not reliable. To address this, we propose enhancing geometric consistency by leveraging temporal cross-view information.

Under the assumption that the depth of static regions remains consistent over time across varying views, we introduce a cross-view temporal-spatial consistency module. For a static pixel $(u_r, v_r)$ in the reference frame with a depth value $d_r$, we project it to the nearest neighboring view—the view with the largest overlap. Using the camera intrinsics $K$ and extrinsics $T_r, T_n$, the corresponding pixel location in the neighboring view is calculated as:
\begin{equation}
[u_n, v_n, 1]^T = K T_n T_r^{-1} \left(d_r \cdot K^{-1}  [u_r, v_r, 1]^T \right).
\end{equation}

We then query the depth value $d_n$ at $(u_n, v_n)$ in the neighboring view. Projecting this back into 3D space, the resulting position should align with the position obtained by back-projecting $(u_r, v_r, d_r)$  to the reference frame:
\begin{equation}
[u_{nr}, v_{nr}, 1]^T = K T_r  T_n^{-1} \left(d_n \cdot  K^{-1}[u_n, v_n, 1]^T \right).
\end{equation}

To enforce cross-view depth consistency, we apply a geometric loss to optimize the Gaussians, defined as:
\begin{equation}
\label{eq:geo-reg}
    \mathcal{L}_{uv} = \| (u_r, v_r) - (u_{nr}, v_{nr}) \|_2.
\end{equation}

This loss encourages the Gaussians to produce geometrically consistent depth across views over time.

\subsection{Optimization}
\noindent{}{\bf Stage I:} During Stage I, our objective is to leverage the joint optimization of motion masks and rendered images to effectively learn the motion masks. Therefore, we only use the masked image losses $\mathcal{L}_I$,
\begin{equation}
   \mathcal{L}_I = (1- \lambda_{ssim})\|I - \tilde{I}\|_1 + \lambda_{ssim}  \texttt{SSIM}(I,\tilde{I}).
\end{equation}
Combined with the motion loss from Eq.\ref{eq:L-d}, we can get:
\begin{equation}
    \mathcal{L}_{stage1} = M \odot \mathcal{L}_I + \mathcal{L}_{dyn}.
\end{equation}

\vspace{5pt}
\noindent{}{\bf Stage II:}
We use the alpha-blending to render the depth map, normal map and velocity map as follows:
\begin{equation}
    \{\mathcal{D},\mathcal{N}, \mathcal{V}\} = \sum_{i \in N} \alpha_i \prod_{j=1}^{i-1} (1 - \alpha_j)\{d_i, n_i, v_i\}.
\end{equation}

For stage II, we use the projected sparse depth map $D_{gt}$ from LiDAR as the supervision label.
\begin{equation}
    \mathcal{L}_{D} = \| \mathcal{D} - D_{gt} \|_1.
\end{equation}

Together with the static velocity regularization (Eq. \ref{eq:vreg}), flattened gaussian (Eq. \ref{eq:min-scale-reg}), normal supervision (Eq. \ref{eq:normal}), giant gaussian regularization (Eq. \ref{eq:big-scale-reg}), geometric consistency loss (Eq. \ref{eq:geo-reg}), etc., the loss for stage II is:
\begin{equation}
  \mathcal{L}_{stage2} = \mathcal{L}_I + \mathcal{L}_D + \mathcal{L}_n + \mathcal{L}_v + \mathcal{L}_s + \mathcal{L}_g + \mathcal{L}_{uv}.
\end{equation}

\section{Experiments}

\begin{table*}[t]
    \centering
    \setlength{\tabcolsep}{2pt}
    \resizebox{\textwidth}{!}{%
    \begin{tabular}{lcccccccccccccc}
        \toprule
       \multirow{3}{*}{\textbf{Method}}  & \multicolumn{7}{c}{\textbf{Waymo Open Dataset}} & \multicolumn{7}{c}{\textbf{KITTI}} \\
    \cmidrule(lr){2-8} \cmidrule(lr){9-15} 
         & \multicolumn{3}{c}{{Image reconstruction}} & \multicolumn{3}{c}{{Novel view synthesis}} &  \multirow{2}{*}{{FPS}} & \multicolumn{3}{c}{{Image reconstruction}} & \multicolumn{3}{c}{{Novel view synthesis}} & \multirow{2}{*}{{FPS}} \\
        \cmidrule(lr){2-4} \cmidrule(lr){5-7}  \cmidrule(lr){9-11} \cmidrule(lr){12-14} 
         & PSNR $\uparrow$ & SSIM $\uparrow$ & LPIPS $\downarrow$ & PSNR $\uparrow$ & SSIM $\uparrow$ & LPIPS $\downarrow$ & & PSNR $\uparrow$ & SSIM $\uparrow$ & LPIPS $\downarrow$ & PSNR $\uparrow$ & SSIM $\uparrow$ & LPIPS $\downarrow$ & \\
        \midrule
        S-NeRF~\cite{snerf_2023_arxiv} & 19.67 & 0.528 & 0.387 & 19.22 & 0.515 & 0.400 & 0.0014 & 19.23 & 0.664 & 0.193 & 18.71 & 0.606 & 0.352 & 0.0075 \\
        StreetSurf~\cite{streetsurf_2023_arxiv} & 26.70 & 0.846 & 0.3717 & 23.78 & 0.822 & 0.401 & 0.097 & 24.14 & 0.819 & 0.257 & 22.48 & 0.763 & 0.304 & 0.37 \\
        3DGS~\cite{3dgs_2023_TOG} & 27.99 & 0.866 & 0.293 & 25.08 & 0.822 & 0.319 & 63 & 21.02 & 0.811 & 0.202 & 19.54 & 0.776 & 0.224 & 125 \\
        NSG~\cite{nsg_2021_cvpr} & 24.08 & 0.656 & 0.441 & 21.01 & 0.571 & 0.487 & 0.032 & 19.19 & 0.683 & 0.189 & 17.78 & 0.645 & 0.312 & 0.19 \\
        Mars~\cite{mars_2023_caai} & 21.81 & 0.681 & 0.430 & 20.69 & 0.636 & 0.453 & 0.030 & 27.96 & 0.900 & 0.185 & 24.31 & 0.845 & 0.160 & 0.31 \\
        SUDS~\cite{suds_2023_cvpr} & 28.83 & 0.805 & 0.317 & 25.36 & 0.783 & 0.384 & 0.008 & 28.83 & 0.917 & 0.147 & 26.07 & 0.797 & 0.131 & 0.29 \\
        EmerNeRF~\cite{emernerf_2023_arxiv} & 28.11 & 0.786 & 0.373 & 25.92 & 0.763 & 0.384 & 0.053 & 26.95 & 0.828 & 0.218 & 25.24 & 0.801 & 0.237 & 0.28 \\
        PVG \cite{pvg_2023_arxiv} & {32.46} & {0.910} & {0.229} & {28.11} & {0.849} & {0.279} & 50 & {32.83} & {0.937} & {0.070} & {27.43} & {0.896} & {0.114} & 59 \\ 
        \midrule
        Ours & \textbf{33.61} & \textbf{0.919} & \textbf{0.204} & \textbf{29.75} & \textbf{0.878} & \textbf{0.213} & 36 & \textbf{33.94} & \textbf{0.949} & \textbf{0.04} & \textbf{28.87} & \textbf{0.901} & \textbf{0.106} & 41 \\
        \bottomrule
    \end{tabular}
    }
    \caption{Comparison of methods on the Waymo Open Dataset and KITTI dataset. FPS refers to frames per second.}
    \label{tab:results}
\end{table*}

\begin{table}[htbp]
\centering
    \resizebox{\linewidth}{!}{%
\begin{tabular}{lc|cc}
\toprule
\textbf{Methods} & \textbf{Box} & \textbf{PSNR (reconst)} $\uparrow$ & \textbf{PSNR (nvs)} $\uparrow$\\
\midrule
EmerNeRF \cite{emernerf_2023_arxiv}&   & 31.93 & 29.67\\
3DGS \cite{emernerf_2023_arxiv}      & & 26.00 & 25.57\\
DeformGS \cite{deformgs_2024_cvpr}  &  & 28.40 & 27.72\\
PVG \cite{pvg_2023_arxiv}        &  & 32.37 & 30.19\\
\midrule
HUGS \cite{hugs_2024_cvpr}      &  \checkmark  & 28.26 & 27.65\\
StreetGS \cite{streetgs_2024_eccv}  &  \checkmark & 29.08 & 28.54\\
OmniRe \cite{omnire_2024_arxiv}   & \checkmark & \textbf{34.25}& \textbf{32.57}\\
\midrule
\textbf{Ours}      &  & \underline{33.82}& \underline{31.49}\\
\bottomrule
\end{tabular}}
\caption{Comparison of rendering quality against recent SOTA methods with or without 3D bbox annotations. `reconst' refers to reconstruction and `nvs' refers to novel view synthesis.}
\label{tab:omnire}
\vspace{-10pt}
\end{table}

\subsection{Experimental Setups}

\noindent \textbf{Datasets.} We conduct our experiments on the Waymo Open Dataset \cite{waymo_2020_cvpr} and KITTI Dataset \cite{kitti_2012_cvpr}, both consisting of real-world autonomous driving scenarios. For the Waymo Open Dataset, we use the subset from PVG \cite{pvg_2023_arxiv}. For a more complete comparison with non self-supervised methods, we also conduct experiments on the subset provided by OmniRe \cite{omnire_2024_arxiv}, which contains a large amount of highly dynamic scenes. We use the frontal three cameras (FRONT LEFT, FRONT, FRONT RIGHT) for Waymo Open Dataset, and the left and right cameras for KITTI dataset.



\begin{table*}[t]
    \centering
    \begin{tabular}{lcccccc}
        \toprule
        Setting
         & PSNR $\uparrow$ & SSIM $\uparrow$ & LPIPS $\downarrow$ & DPSNR $\uparrow$ & DSSIM $\uparrow$ & Depth L1$\downarrow$
         \\
        \midrule
       (a) w/o stage I motion mask & 34.7063 & 0.9570 & 0.1098 & 34.7183 & 0.9570 & 0.1017 \\
       (b) w/o FiT3D model (w DINOv2) & 34.9551 & 0.9559 & 0.1027 & 34.9734 & 0.9602 & 0.0977 \\
       (c) w/o gt normal supervision & 35.4469 & 0.9625 & 0.0967 & 35.4876 & 0.9626 & 0.0913 \\
       (d) w/o gt normal (w normal from depth)  & 35.2357 & 0.9509 & 0.1436 & 35.5312 & 0.9512 &  0.0847 \\
       (e) w/o min scale regularization~ & 35.2863 & 0.9616 & 0.0989 & 35.3275 & 0.9617 & 0.0935 \\
       (f) w/o max scale regularization~ & 35.6911 & 0.9622 & 0.0970 & 35.7306 & 0.9623 & 0.0802 \\
       (g) w/o multi-view consistency  & 35.3325 & 0.9618 & 0.0983 & 35.3731 & 0.9619 & 0.1154 \\
        \midrule
        Full model  & \textbf{35.7598} & \textbf{0.9631} & \textbf{0.0956} & \textbf{35.7820} & \textbf{0.9632} & \textbf{0.0713} \\
        \bottomrule
    \end{tabular}
    \caption{Ablations Studies.}
    \label{tab:ablations}
    \vspace{-10pt}
\end{table*}

\begin{figure*}[th!]
    \centering
    \includegraphics[width=0.92\linewidth]{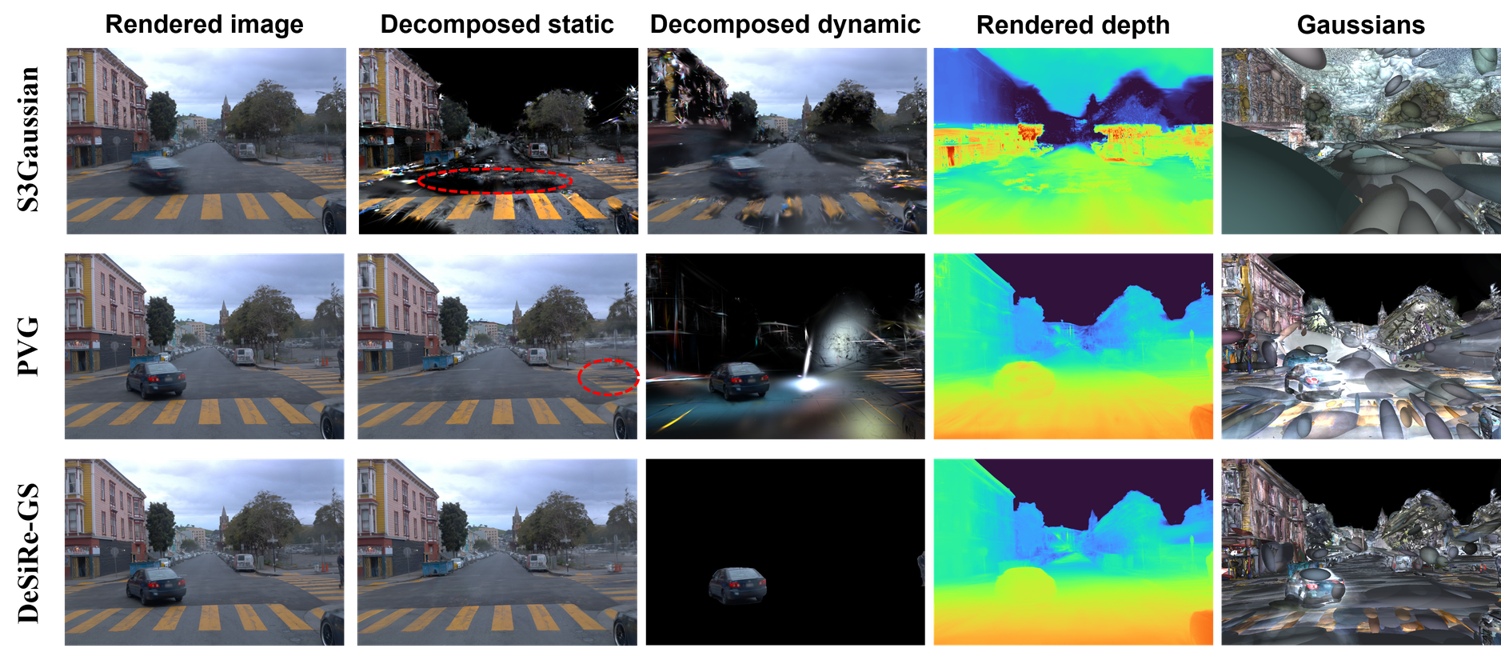}
    \caption{Qualitative comparison with self-supervised S3Gaussian \cite{s3g_2024_arxiv} and PVG \cite{pvg_2023_arxiv}}
    \label{fig:qualitative-comparison}
\end{figure*}

\vspace{5pt}
\noindent \textbf{Evaluation Metrics.} 
We adopt PSNR, SSIM \cite{ssim_2004} and LPIPS \cite{lpips_cvpr_2018} as metrics for the evaluation of image reconstruction and novel view synthesis. Following \cite{emernerf_2023_arxiv,streetgs_2024_eccv,s3g_2024_arxiv}, we also include DPSNR and DSSIM to assess the rendering quality at dynamic regions. Additionally, we introduce depth L1, which measures the L1 error between the rendered depth map and the ground truth depth map obtained from LiDAR point clouds, as an evaluation metric for the quality of geometric reconstruction.

\vspace{5pt}
\noindent \textbf{Baselines.} We benchmark \ourmethod against the following approaches: 3DGS \cite{3dgs_2023_TOG}, StreetSurf \cite{streetsurf_2023_arxiv}, Mars \cite{mars_2023_caai}, SUDS \cite{suds_2023_cvpr}, EmerNeRF \cite{emernerf_2023_arxiv}, S3Gaussian \cite{s3g_2024_arxiv}, PVG \cite{pvg_2023_arxiv}, OmniRe \cite{omnire_2024_arxiv}, StreetGS \cite{streetgs_2024_eccv} and HUGS\cite{hugs_2024_cvpr}. Among these methods, SUDS and EmerNeRF are NeRF-based self-supervised approaches. S3Gaussian and PVG are both 3DGS-based self-supervised methods, the closest to our approach. To further highlight the superiority of DeSiRe-GS, we also compare it with OmniRe, StreetGS, and HUGS, all of which require additional bounding box information.



\vspace{5pt}
\noindent \textbf{Implementation Details.}  All experiments are conducted on NVIDIA RTX A6000. We sample a total of \(1 \times 10^6\) points for initialization, with \(6 \times 10^5\) from LiDAR point cloud, and \(4 \times 10^5\) randomly sampled points. In the first stage, we train for a total of 30,000 iterations. We start to train the motion decoder after 6,000 iterations. For the second stage, we train the model for 50,000 iterations. Multi-view temporal consistency regularization begins after 20,000 iterations. The motion masks, obtained from stage I, are employed after 30000 iterations to supervise the optimization of velocity \(\boldsymbol{v}\).  We use Adam \cite{adam_iclr_2015} as our optimizer with $\beta_1 = 0.9, \beta_2 = 0.999$.


\subsection{Quantitative Results}
Following PVG \cite{pvg_2023_arxiv}, we evaluate our method on two tasks: image reconstruction and novel view synthesis, using the Waymo Open Dataset \cite{waymo_2020_cvpr} and the KITTI dataset \cite{kitti_2012_cvpr}. As shown in Tab. \ref{tab:results}, our approach achieves state-of-the-art performance across all rendering metrics for both reconstruction and synthesis tasks. In terms of rendering speed, our method reaches approximately 40 FPS. While slightly slower than the 3DGS \cite{3dgs_2023_TOG} and PVG \cite{pvg_2023_arxiv} baselines due to rendering additional attributes such as normals, our approach delivers over a 1.1 PSNR improvement.

In addition to comparisons with self-supervised methods, we evaluate against approaches that rely on 3D annotations. The results, detailed in Table \ref{tab:omnire}, show that our method achieves comparable, if not superior, performance to baselines like HUGS \cite{hugs_2024_cvpr} and StreetGS \cite{streetgs_2024_eccv}, while outperforming self-supervised baselines \cite{emernerf_2023_arxiv, pvg_2023_arxiv}.

\begin{figure}
    \centering
    \includegraphics[width=\linewidth]{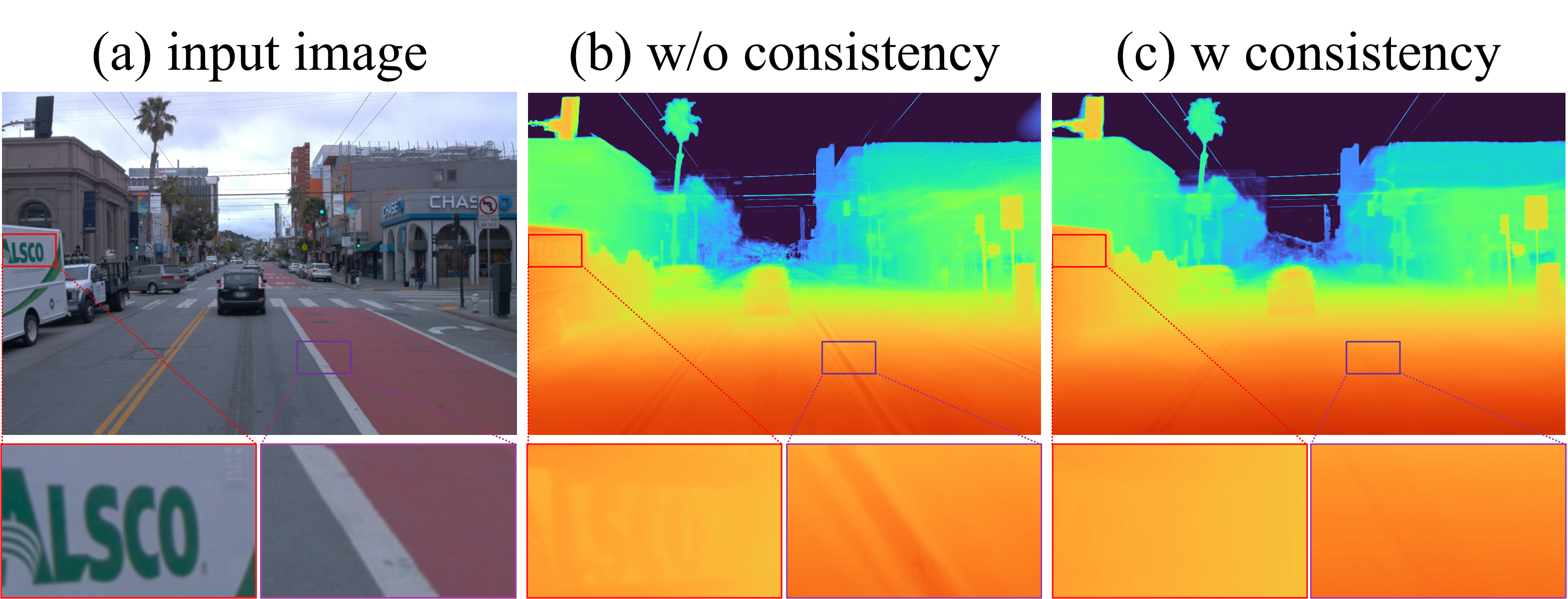}
    \caption{Multi-view consistency depth (Better viewed zoom-in)}
    \label{fig:multi_view_consistency_depth}
    \vspace{-10pt}
\end{figure}

\subsection{Qualitative Analysis}
We provide visualizations of static-dynamic decomposition and depth prediction in Fig. \ref{fig:qualitative-comparison}. S3Gaussian \cite{s3g_2024_arxiv} fails to generate satisfactory depth maps or achieve effective static-dynamic decomposition. Similarly, PVG \cite{pvg_2023_arxiv} produces only blurry and suboptimal decomposition results.

\subsection{Ablation Studies}

To verify the effectiveness of our proposed components, we conduct ablation studies on Waymo Open Dataset. The results are listed in Tab. \ref{tab:ablations}.

\vspace{5pt}
\noindent \textbf{Motion mask.} We train our model from scratch at stage II without using the motion masks obtained from stage I for regularization. As shown in Tab.\ref{tab:ablations} (a), the motion masks from Stage I improve both the rendering and reconstruction quality by a large margin.

We also ablate on different foundation models as the feature extractor. As shown in Tab.\ref{tab:ablations} (b), the FiT3D model \cite{fit3d_2024_eccv} outperforms DINOv2, producing much better results.

\vspace{5pt}
\noindent \textbf{Normal Supervision.}
For normal supervision, we explored two approaches. The first approach utilizes powerful pre-trained models, such as OmniData \cite{omnidata_2021_iccv}, to predict pseudo-normal maps $\hat{\mathcal{N}}$ from the input monocular images. The second approach employs the depth gradient as the pseudo-normal map $\hat{\mathcal{N}}_D$ for supervision \cite{gsir_2024_cvpr}. 

As shown in Tab. \ref{tab:ablations} (c)(d), we found that pseudo-normals predicted from Omnidata produce the best overall results. While using $\hat{\mathcal{N}}_D$ slightly improves depth accuracy, it leads to suboptimal rendering quality. We attribute this to the reliance on sparse depth maps projected from LiDAR point clouds for supervision. Although our rendered depth maps and the corresponding normal maps (derived from depth gradients) are dense, the supervision remains incomplete due to the inherent sparsity of the LiDAR points.

\vspace{5pt}
\noindent \textbf{Scale Regularization.}
We also impose constraints on the size of the Gaussians, ensuring their scale remains within a reasonable range. As shown in Tab. \ref{tab:ablations} (e)(f), the improvements in rendering metrics are not particularly significant. We attribute this to the strong overfitting capability of Gaussians. Despite the presence of some oversized Gaussian outliers, the final rendering results remain visually satisfactory. However, as illustrated in Figure \ref{fig:gaint_reg}, the Gaussians produced with regularization exhibit improved 3D structure, enabling more effective decomposition.

\vspace{5pt}
\noindent \textbf{Cross-view Consistency.}
As shown in Tab. \ref{tab:ablations} (g), the proposed cross-view consistency significantly enhances the geometric metrics. Fig. \ref{fig:multi_view_consistency_depth} demonstrates that, without our method, Gaussians tend to overfit to textured areas in the image, such as the slogan and white line, resulting in unexpected large depth variations. With our multi-view consistency module, this overfitting issue is effectively mitigated.
\section{Conclusion}
In this paper, we propose \ourmethod, a self-supervised approach for static-dynamic decomposition and high-quality surface reconstruction in driving scenes. By introducing a motion mask module and leveraging temporal geometrical consistency, \ourmethod addresses key challenges such as dynamic object modeling and data sparsity.


\noindent \textbf{Achknowledgements}\\
This work was supported by Berkeley DeepDrive. \footnote{\url{https://deepdrive.berkeley.edu}} 
{
    \small
    \bibliographystyle{ieeenat_fullname}
    \bibliography{main}

\begin{thebibliography}{53}
\providecommand{\natexlab}[1]{#1}
\providecommand{\url}[1]{\texttt{#1}}
\expandafter\ifx\csname urlstyle\endcsname\relax
  \providecommand{\doi}[1]{doi: #1}\else
  \providecommand{\doi}{doi: \begingroup \urlstyle{rm}\Url}\fi

\bibitem[Cao and Johnson(2023)]{hexplane_2023_cvpr}
Ang Cao and Justin Johnson.
\newblock Hexplane: A fast representation for dynamic scenes.
\newblock In \emph{CVPR}, pages 130--141, 2023.

\bibitem[Chen et~al.(2024{\natexlab{a}})Chen, Li, Ye, Wang, Xie, Zhai, Wang, Liu, Bao, and Zhang]{pgsr_2024_arxiv}
Danpeng Chen, Hai Li, Weicai Ye, Yifan Wang, Weijian Xie, Shangjin Zhai, Nan Wang, Haomin Liu, Hujun Bao, and Guofeng Zhang.
\newblock Pgsr: Planar-based gaussian splatting for efficient and high-fidelity surface reconstruction.
\newblock \emph{arXiv preprint arXiv:2406.06521}, 2024{\natexlab{a}}.

\bibitem[Chen et~al.(2023{\natexlab{a}})Chen, Li, and Lee]{neusg_2023_cvpr}
Hanlin Chen, Chen Li, and Gim~Hee Lee.
\newblock Neusg: Neural implicit surface reconstruction with 3d gaussian splatting guidance, 2023{\natexlab{a}}.

\bibitem[Chen et~al.(2024{\natexlab{b}})Chen, Ye, Wang, Chen, Huang, Ouyang, Zhang, Qiao, and He]{gigags_2024_arxiv}
Junyi Chen, Weicai Ye, Yifan Wang, Danpeng Chen, Di Huang, Wanli Ouyang, Guofeng Zhang, Yu Qiao, and Tong He.
\newblock Gigags: Scaling up planar-based 3d gaussians for large scene surface reconstruction.
\newblock \emph{arXiv preprint arXiv:2409.06685}, 2024{\natexlab{b}}.

\bibitem[Chen et~al.(2023{\natexlab{b}})Chen, Gu, Jiang, Zhu, and Zhang]{pvg_2023_arxiv}
Yurui Chen, Chun Gu, Junzhe Jiang, Xiatian Zhu, and Li Zhang.
\newblock Periodic vibration gaussian: Dynamic urban scene reconstruction and real-time rendering.
\newblock \emph{arXiv preprint arXiv:2311.18561}, 2023{\natexlab{b}}.

\bibitem[Chen et~al.(2024{\natexlab{c}})Chen, Yang, Huang, de~Lutio, Esturo, Ivanovic, Litany, Gojcic, Fidler, Pavone, et~al.]{omnire_2024_arxiv}
Ziyu Chen, Jiawei Yang, Jiahui Huang, Riccardo de Lutio, Janick~Martinez Esturo, Boris Ivanovic, Or Litany, Zan Gojcic, Sanja Fidler, Marco Pavone, et~al.
\newblock Omnire: Omni urban scene reconstruction.
\newblock \emph{arXiv preprint arXiv:2408.16760}, 2024{\natexlab{c}}.

\bibitem[Cheng et~al.(2024)Cheng, Long, Yang, Yao, Yin, Ma, Wang, and Chen]{gaussianpro_2024_icml}
Kai Cheng, Xiaoxiao Long, Kaizhi Yang, Yao Yao, Wei Yin, Yuexin Ma, Wenping Wang, and Xuejin Chen.
\newblock Gaussianpro: 3d gaussian splatting with progressive propagation.
\newblock In \emph{Forty-first International Conference on Machine Learning}, 2024.

\bibitem[Eftekhar et~al.(2021)Eftekhar, Sax, Malik, and Zamir]{omnidata_2021_iccv}
Ainaz Eftekhar, Alexander Sax, Jitendra Malik, and Amir Zamir.
\newblock Omnidata: A scalable pipeline for making multi-task mid-level vision datasets from 3d scans.
\newblock In \emph{ICCV}, pages 10786--10796, 2021.

\bibitem[Fan et~al.(2024)Fan, Yang, Li, Li, and Zhang]{trim3d_2024_cvpr}
Lue Fan, Yuxue Yang, Minxing Li, Hongsheng Li, and Zhaoxiang Zhang.
\newblock Trim 3d gaussian splatting for accurate geometry representation, 2024.

\bibitem[Feng et~al.(2024)Feng, Wu, and Wang]{rogs_2024_cvpr}
Zhiheng Feng, Wenhua Wu, and Hesheng Wang.
\newblock Rogs: Large scale road surface reconstruction based on 2d gaussian splatting, 2024.

\bibitem[Geiger et~al.(2012)Geiger, Lenz, and Urtasun]{kitti_2012_cvpr}
Andreas Geiger, Philip Lenz, and Raquel Urtasun.
\newblock Are we ready for autonomous driving? the kitti vision benchmark suite.
\newblock In \emph{Conference on Computer Vision and Pattern Recognition (CVPR)}, 2012.

\bibitem[Gu{\'e}don and Lepetit(2024)]{sugar_2024_cvpr}
Antoine Gu{\'e}don and Vincent Lepetit.
\newblock Sugar: Surface-aligned gaussian splatting for efficient 3d mesh reconstruction and high-quality mesh rendering.
\newblock In \emph{CVPR}, pages 5354--5363, 2024.

\bibitem[Guo et~al.(2023)Guo, Deng, Li, Bai, Shi, Wang, Ding, Wang, and Li]{streetsurf_2023_arxiv}
Jianfei Guo, Nianchen Deng, Xinyang Li, Yeqi Bai, Botian Shi, Chiyu Wang, Chenjing Ding, Dongliang Wang, and Yikang Li.
\newblock Streetsurf: Extending multi-view implicit surface reconstruction to street views.
\newblock \emph{arXiv preprint arXiv:2306.04988}, 2023.

\bibitem[Huang et~al.(2024{\natexlab{a}})Huang, Yu, Chen, Geiger, and Gao]{2dgs_2024_siggraph}
Binbin Huang, Zehao Yu, Anpei Chen, Andreas Geiger, and Shenghua Gao.
\newblock 2d gaussian splatting for geometrically accurate radiance fields.
\newblock In \emph{ACM SIGGRAPH 2024 Conference Papers}, pages 1--11, 2024{\natexlab{a}}.

\bibitem[Huang et~al.(2024{\natexlab{b}})Huang, Wei, Zheng, An, Lu, Zhan, Tomizuka, Keutzer, and Zhang]{s3g_2024_arxiv}
Nan Huang, Xiaobao Wei, Wenzhao Zheng, Pengju An, Ming Lu, Wei Zhan, Masayoshi Tomizuka, Kurt Keutzer, and Shanghang Zhang.
\newblock S3gaussian: Self-supervised street gaussians for autonomous driving.
\newblock \emph{CoRR}, 2024{\natexlab{b}}.

\bibitem[Kerbl et~al.(2023)Kerbl, Kopanas, Leimk{\"u}hler, and Drettakis]{3dgs_2023_TOG}
Bernhard Kerbl, Georgios Kopanas, Thomas Leimk{\"u}hler, and George Drettakis.
\newblock 3d gaussian splatting for real-time radiance field rendering.
\newblock \emph{ACM Trans. Graph.}, 42\penalty0 (4):\penalty0 139--1, 2023.

\bibitem[Kingma and Ba(2017)]{adam_iclr_2015}
Diederik~P. Kingma and Jimmy Ba.
\newblock Adam: A method for stochastic optimization, 2017.

\bibitem[Kulhanek et~al.(2024)Kulhanek, Peng, Kukelova, Pollefeys, and Sattler]{wildgaussians_2024_nips}
Jonas Kulhanek, Songyou Peng, Zuzana Kukelova, Marc Pollefeys, and Torsten Sattler.
\newblock {W}ild{G}aussians: {3D} gaussian splatting in the wild.
\newblock \emph{arXiv}, 2024.

\bibitem[Li et~al.(2023)Li, Müller, Evans, Taylor, Unberath, Liu, and Lin]{neuralangelo_2023_cvpr}
Zhaoshuo Li, Thomas Müller, Alex Evans, Russell~H. Taylor, Mathias Unberath, Ming-Yu Liu, and Chen-Hsuan Lin.
\newblock Neuralangelo: High-fidelity neural surface reconstruction, 2023.

\bibitem[Liang et~al.(2024)Liang, Zhang, Feng, Shan, and Jia]{gsir_2024_cvpr}
Zhihao Liang, Qi Zhang, Ying Feng, Ying Shan, and Kui Jia.
\newblock Gs-ir: 3d gaussian splatting for inverse rendering.
\newblock In \emph{CVPR}, pages 21644--21653, 2024.

\bibitem[Liu et~al.(2024{\natexlab{a}})Liu, Yu, Wang, Zheng, Deng, Ye, and Wang]{liu2024point}
Jiuming Liu, Ruiji Yu, Yian Wang, Yu Zheng, Tianchen Deng, Weicai Ye, and Hesheng Wang.
\newblock Point mamba: A novel point cloud backbone based on state space model with octree-based ordering strategy.
\newblock \emph{arXiv preprint arXiv:2403.06467}, 2024{\natexlab{a}}.

\bibitem[Liu et~al.(2024{\natexlab{b}})Liu, Zhuo, Feng, Zhu, Peng, Liu, and Wang]{liu2024dvlo}
Jiuming Liu, Dong Zhuo, Zhiheng Feng, Siting Zhu, Chensheng Peng, Zhe Liu, and Hesheng Wang.
\newblock Dvlo: Deep visual-lidar odometry with local-to-global feature fusion and bi-directional structure alignment.
\newblock In \emph{European Conference on Computer Vision}, pages 475--493. Springer, 2024{\natexlab{b}}.

\bibitem[Mildenhall et~al.(2020)Mildenhall, Srinivasan, Tancik, Barron, Ramamoorthi, and Ng]{nerf_2020_eccv}
Ben Mildenhall, Pratul~P. Srinivasan, Matthew Tancik, Jonathan~T. Barron, Ravi Ramamoorthi, and Ren Ng.
\newblock Nerf: Representing scenes as neural radiance fields for view synthesis, 2020.

\bibitem[Nguyen et~al.(2024)Nguyen, Rold{\~a}o, Piasco, Bennehar, and Tsishkou]{rodus_2024_eccv}
Thang-Anh-Quan Nguyen, Luis Rold{\~a}o, Nathan Piasco, Moussab Bennehar, and Dzmitry Tsishkou.
\newblock Rodus: Robust decomposition of static and dynamic elements in urban scenes.
\newblock \emph{arXiv preprint arXiv:2403.09419}, 2024.

\bibitem[Oquab et~al.(2023)Oquab, Darcet, Moutakanni, Vo, Szafraniec, Khalidov, Fernandez, Haziza, Massa, El-Nouby, et~al.]{dinov2_2023_arxiv}
Maxime Oquab, Timoth{\'e}e Darcet, Th{\'e}o Moutakanni, Huy Vo, Marc Szafraniec, Vasil Khalidov, Pierre Fernandez, Daniel Haziza, Francisco Massa, Alaaeldin El-Nouby, et~al.
\newblock Dinov2: Learning robust visual features without supervision.
\newblock \emph{arXiv preprint arXiv:2304.07193}, 2023.

\bibitem[Ost et~al.(2021)Ost, Mannan, Thuerey, Knodt, and Heide]{nsg_2021_cvpr}
Julian Ost, Fahim Mannan, Nils Thuerey, Julian Knodt, and Felix Heide.
\newblock Neural scene graphs for dynamic scenes.
\newblock In \emph{CVPR}, pages 2856--2865, 2021.

\bibitem[Park et~al.(2021)Park, Sinha, Barron, Bouaziz, Goldman, Seitz, and Martin-Brualla]{nerfiles_iccv_2021}
Keunhong Park, Utkarsh Sinha, Jonathan~T. Barron, Sofien Bouaziz, Dan~B Goldman, Steven~M. Seitz, and Ricardo Martin-Brualla.
\newblock Nerfies: Deformable neural radiance fields, 2021.

\bibitem[Peng et~al.(2023)Peng, Wang, Lo, Wu, Xu, Tomizuka, Zhan, and Wang]{peng2023delflow}
Chensheng Peng, Guangming Wang, Xian~Wan Lo, Xinrui Wu, Chenfeng Xu, Masayoshi Tomizuka, Wei Zhan, and Hesheng Wang.
\newblock Delflow: Dense efficient learning of scene flow for large-scale point clouds.
\newblock In \emph{ICCV}, pages 16901--16910, 2023.

\bibitem[Peng et~al.(2024{\natexlab{a}})Peng, Xu, Wang, Ding, Yang, Tomizuka, Keutzer, Pavone, and Zhan]{qslam}
Chensheng Peng, Chenfeng Xu, Yue Wang, Mingyu Ding, Heng Yang, Masayoshi Tomizuka, Kurt Keutzer, Marco Pavone, and Wei Zhan.
\newblock Q-slam: Quadric representations for monocular slam.
\newblock \emph{Conference on Robot Learning (CoRL)}, 2024{\natexlab{a}}.

\bibitem[Peng et~al.(2024{\natexlab{b}})Peng, Zeng, Gao, Zhou, Tomizuka, Wang, Zhou, and Ye]{pnasmot}
Chensheng Peng, Zhaoyu Zeng, Jinling Gao, Jundong Zhou, Masayoshi Tomizuka, Xinbing Wang, Chenghu Zhou, and Nanyang Ye.
\newblock Pnas-mot: multi-modal object tracking with pareto neural architecture search.
\newblock \emph{IEEE Robotics and Automation Letters}, 9\penalty0 (5):\penalty0 4377--4384, 2024{\natexlab{b}}.

\bibitem[Pumarola et~al.(2020)Pumarola, Corona, Pons-Moll, and Moreno-Noguer]{dnerf_cvpr_2020}
Albert Pumarola, Enric Corona, Gerard Pons-Moll, and Francesc Moreno-Noguer.
\newblock D-nerf: Neural radiance fields for dynamic scenes, 2020.

\bibitem[Rematas et~al.(2022)Rematas, Liu, Srinivasan, Barron, Tagliasacchi, Funkhouser, and Ferrari]{urbannerf_2022_cvpr}
Konstantinos Rematas, Andrew Liu, Pratul~P Srinivasan, Jonathan~T Barron, Andrea Tagliasacchi, Thomas Funkhouser, and Vittorio Ferrari.
\newblock Urban radiance fields.
\newblock In \emph{CVPR}, pages 12932--12942, 2022.

\bibitem[Sun et~al.(2020)Sun, Kretzschmar, Dotiwalla, Chouard, Patnaik, Tsui, Guo, Zhou, Chai, Caine, et~al.]{waymo_2020_cvpr}
Pei Sun, Henrik Kretzschmar, Xerxes Dotiwalla, Aurelien Chouard, Vijaysai Patnaik, Paul Tsui, James Guo, Yin Zhou, Yuning Chai, Benjamin Caine, et~al.
\newblock Scalability in perception for autonomous driving: Waymo open dataset.
\newblock In \emph{CVPR}, pages 2446--2454, 2020.

\bibitem[Tancik et~al.(2022)Tancik, Casser, Yan, Pradhan, Mildenhall, Srinivasan, Barron, and Kretzschmar]{blocknerf_2022_cvpr}
Matthew Tancik, Vincent Casser, Xinchen Yan, Sabeek Pradhan, Ben Mildenhall, Pratul~P. Srinivasan, Jonathan~T. Barron, and Henrik Kretzschmar.
\newblock Block-nerf: Scalable large scene neural view synthesis, 2022.

\bibitem[Turki et~al.(2022)Turki, Ramanan, and Satyanarayanan]{meganerf_2022_cvpr}
Haithem Turki, Deva Ramanan, and Mahadev Satyanarayanan.
\newblock Mega-nerf: Scalable construction of large-scale nerfs for virtual fly-throughs, 2022.

\bibitem[Turki et~al.(2023)Turki, Zhang, Ferroni, and Ramanan]{suds_2023_cvpr}
Haithem Turki, Jason~Y Zhang, Francesco Ferroni, and Deva Ramanan.
\newblock Suds: Scalable urban dynamic scenes.
\newblock In \emph{CVPR}, pages 12375--12385, 2023.

\bibitem[Turkulainen et~al.(2024)Turkulainen, Ren, Melekhov, Seiskari, Rahtu, and Kannala]{dnsplatter_2024_cvpr}
Matias Turkulainen, Xuqian Ren, Iaroslav Melekhov, Otto Seiskari, Esa Rahtu, and Juho Kannala.
\newblock Dn-splatter: Depth and normal priors for gaussian splatting and meshing, 2024.

\bibitem[Wang et~al.(2023)Wang, Peng, Gu, Zhang, and Wang]{wang2023interactive}
Guangming Wang, Chensheng Peng, Yingying Gu, Jinpeng Zhang, and Hesheng Wang.
\newblock Interactive multi-scale fusion of 2d and 3d features for multi-object vehicle tracking.
\newblock \emph{IEEE Transactions on Intelligent Transportation Systems}, 24\penalty0 (10):\penalty0 10618--10627, 2023.

\bibitem[Wang et~al.(2021)Wang, Liu, Liu, Theobalt, Komura, and Wang]{neus_2021_arxiv}
Peng Wang, Lingjie Liu, Yuan Liu, Christian Theobalt, Taku Komura, and Wenping Wang.
\newblock Neus: Learning neural implicit surfaces by volume rendering for multi-view reconstruction.
\newblock \emph{arXiv preprint arXiv:2106.10689}, 2021.

\bibitem[Wang et~al.(2004)Wang, Bovik, Sheikh, and Simoncelli]{ssim_2004}
Zhou Wang, A.C. Bovik, H.R. Sheikh, and E.P. Simoncelli.
\newblock Image quality assessment: from error visibility to structural similarity.
\newblock \emph{IEEE Transactions on Image Processing}, 13\penalty0 (4):\penalty0 600--612, 2004.

\bibitem[Wu et~al.(2024)Wu, Yi, Fang, Xie, Zhang, Wei, Liu, Tian, and Wang]{4dgs_2024_cvpr}
Guanjun Wu, Taoran Yi, Jiemin Fang, Lingxi Xie, Xiaopeng Zhang, Wei Wei, Wenyu Liu, Qi Tian, and Xinggang Wang.
\newblock 4d gaussian splatting for real-time dynamic scene rendering, 2024.

\bibitem[Wu et~al.(2023)Wu, Liu, Luo, Zhong, Chen, Xiao, Hou, Lou, Chen, Yang, et~al.]{mars_2023_caai}
Zirui Wu, Tianyu Liu, Liyi Luo, Zhide Zhong, Jianteng Chen, Hongmin Xiao, Chao Hou, Haozhe Lou, Yuantao Chen, Runyi Yang, et~al.
\newblock Mars: An instance-aware, modular and realistic simulator for autonomous driving.
\newblock In \emph{CAAI International Conference on Artificial Intelligence}, pages 3--15. Springer, 2023.

\bibitem[Xie et~al.(2023)Xie, Zhang, Li, Zhang, and Zhang]{snerf_2023_arxiv}
Ziyang Xie, Junge Zhang, Wenye Li, Feihu Zhang, and Li Zhang.
\newblock S-nerf: Neural radiance fields for street views.
\newblock \emph{arXiv preprint arXiv:2303.00749}, 2023.

\bibitem[Yan et~al.(2024)Yan, Lin, Zhou, Wang, Sun, Zhan, Lang, Zhou, and Peng]{streetgs_2024_eccv}
Yunzhi Yan, Haotong Lin, Chenxu Zhou, Weijie Wang, Haiyang Sun, Kun Zhan, Xianpeng Lang, Xiaowei Zhou, and Sida Peng.
\newblock Street gaussians for modeling dynamic urban scenes.
\newblock \emph{arXiv preprint arXiv:2401.01339}, 2024.

\bibitem[Yang et~al.(2023)Yang, Ivanovic, Litany, Weng, Kim, Li, Che, Xu, Fidler, Pavone, et~al.]{emernerf_2023_arxiv}
Jiawei Yang, Boris Ivanovic, Or Litany, Xinshuo Weng, Seung~Wook Kim, Boyi Li, Tong Che, Danfei Xu, Sanja Fidler, Marco Pavone, et~al.
\newblock Emernerf: Emergent spatial-temporal scene decomposition via self-supervision.
\newblock \emph{arXiv preprint arXiv:2311.02077}, 2023.

\bibitem[Yang et~al.(2024)Yang, Gao, Zhou, Jiao, Zhang, and Jin]{deformgs_2024_cvpr}
Ziyi Yang, Xinyu Gao, Wen Zhou, Shaohui Jiao, Yuqing Zhang, and Xiaogang Jin.
\newblock Deformable 3d gaussians for high-fidelity monocular dynamic scene reconstruction.
\newblock In \emph{CVPR}, pages 20331--20341, 2024.

\bibitem[Yariv et~al.(2021)Yariv, Gu, Kasten, and Lipman]{volsdf_2021_cvpr}
Lior Yariv, Jiatao Gu, Yoni Kasten, and Yaron Lipman.
\newblock Volume rendering of neural implicit surfaces, 2021.

\bibitem[Yu et~al.(2024)Yu, Lu, Xu, Jiang, Xiangli, and Dai]{gsdf_2024_cvpr}
Mulin Yu, Tao Lu, Linning Xu, Lihan Jiang, Yuanbo Xiangli, and Bo Dai.
\newblock Gsdf: 3dgs meets sdf for improved rendering and reconstruction, 2024.

\bibitem[Yu et~al.(2022)Yu, Peng, Niemeyer, Sattler, and Geiger]{monosdf_2022_cvpr}
Zehao Yu, Songyou Peng, Michael Niemeyer, Torsten Sattler, and Andreas Geiger.
\newblock Monosdf: Exploring monocular geometric cues for neural implicit surface reconstruction, 2022.

\bibitem[Yue et~al.(2024)Yue, Das, Engelmann, Tang, and Lenssen]{fit3d_2024_eccv}
Yuanwen Yue, Anurag Das, Francis Engelmann, Siyu Tang, and Jan~Eric Lenssen.
\newblock Improving 2d feature representations by 3d-aware fine-tuning.
\newblock In \emph{ECCV}, 2024.

\bibitem[Zhang et~al.(2018)Zhang, Isola, Efros, Shechtman, and Wang]{lpips_cvpr_2018}
Richard Zhang, Phillip Isola, Alexei~A. Efros, Eli Shechtman, and Oliver Wang.
\newblock The unreasonable effectiveness of deep features as a perceptual metric, 2018.

\bibitem[Zhou et~al.(2024{\natexlab{a}})Zhou, Shao, Xu, Bai, Qiu, Liu, Wang, Geiger, and Liao]{hugs_2024_cvpr}
Hongyu Zhou, Jiahao Shao, Lu Xu, Dongfeng Bai, Weichao Qiu, Bingbing Liu, Yue Wang, Andreas Geiger, and Yiyi Liao.
\newblock Hugs: Holistic urban 3d scene understanding via gaussian splatting.
\newblock In \emph{CVPR}, pages 21336--21345, 2024{\natexlab{a}}.

\bibitem[Zhou et~al.(2024{\natexlab{b}})Zhou, Lin, Shan, Wang, Sun, and Yang]{drivinggaussian_2024_cvpr}
Xiaoyu Zhou, Zhiwei Lin, Xiaojun Shan, Yongtao Wang, Deqing Sun, and Ming-Hsuan Yang.
\newblock Drivinggaussian: Composite gaussian splatting for surrounding dynamic autonomous driving scenes.
\newblock In \emph{CVPR}, pages 21634--21643, 2024{\natexlab{b}}.

\end{thebibliography}
}

\clearpage
\setcounter{page}{1}

\twocolumn[{%
\renewcommand\twocolumn[1][]{#1}%
\maketitlesupplementary
\begin{center}
    \centering
    \captionsetup{type=figure}
    \includegraphics[width=\textwidth]{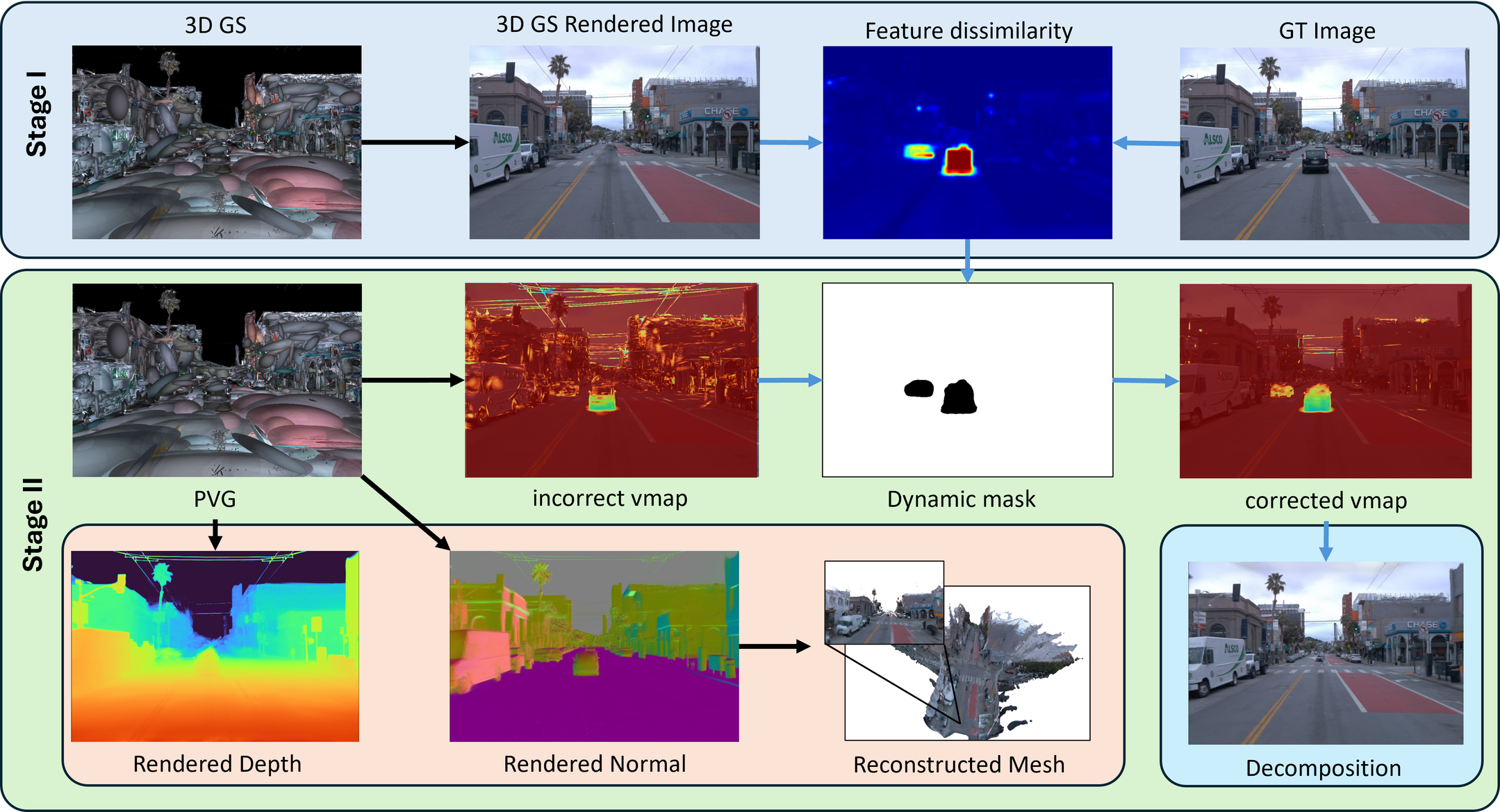}
    \captionof{figure}{\textbf{\ourmethod}. We present a 4D street gaussian splatting representation for self-supervised static-dynamic decomposition and high-fidelity surface reconstruction without the requirement for extra 3D annotations such as bounding boxes.}
    \label{fig:sup_teaser}
\end{center}%
}]


\section{Implementation Details}

All experiments are conducted on NVIDIA RTX A6000. We sample a total of \(1 \times 10^6\) points for initialization, with \(6 \times 10^5\) from LiDAR point cloud, \(2 \times 10^5\) near points and \(2 \times 10^5\) far points depending on their distance to LiDAR origin. In the first stage, we train for a total of 30,000 iterations. We do not train the uncertainty model during the initial 6,000 iterations. After that, the uncertainty model gradually increases its weight over a 1,800-iteration warm-up process. In the second stage, we train a total of 50000 iterations. Cross-view consistency regularization begins after 20,000 iterations, with 102400 pixels sampled each time. Motion masks, obtained from the trained dynamic model, are employed after 30000 iterations to supervise the training of velocity \(\mathcal{\textbf{v}}\) and time scale \(\beta \).  We use Adam \cite{adam_iclr_2015} as our optimizer with $\beta_1 = 0.9, \beta_2 = 0.999$ and maintain similar optimization settings of \cite{pvg_2023_arxiv}. For the dynamic model we employ a learning rate of 0.001 and a dropout rate of 0.1.

\vspace{5pt}
\noindent \textbf{Evaluation Metrics.}
 We adopt Peak Signal-to-Noise Ratio (PSNR), Structural Similarity Index Measure (SSIM) \cite{ssim_2004} and Learned Perceptual Image Patch Similarity (LPIPS) \cite{lpips_cvpr_2018} as metrics for the assessment of both image reconstruction and novel view synthesis. Following \cite{emernerf_2023_arxiv,streetgs_2024_eccv,s3g_2024_arxiv}, we also include DPSNR and DSSIM to assess the rendering quality of dynamic objects. Specifically, these values are calculated by projecting the 3D bounding box of moving objects onto the camera plane and computing the PSNR and SSIM within the projected box. Additionally, we introduce depth L1, which measures the L1 error between the rendered depth map and the ground truth depth map, as an evaluation metric which is related to surface reconstruction.

\begin{figure*}[th!]
    \centering
    \includegraphics[width=\linewidth]{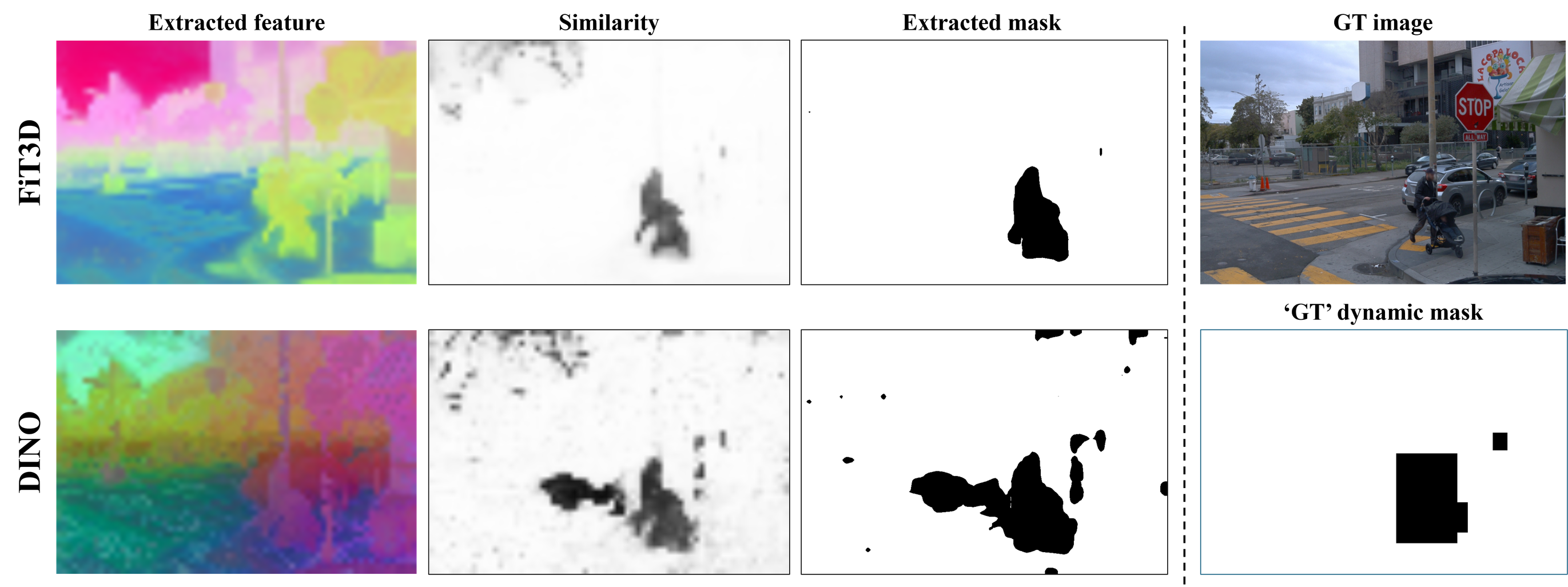}
    \caption{Comparison between  motion masks and features extracted from DINOv2 \cite{dinov2_2023_arxiv} and FiT3D~\cite{fit3d_2024_eccv}.  The `GT' masks are obtained using the GT bounding boxes and associated trajectories following EmerNeRF~\cite{emernerf_2023_arxiv}.}
    \label{fig:dino-vs-fit3d}
\end{figure*} 
\begin{figure*}[th!]
    \centering
    \includegraphics[width=\linewidth]{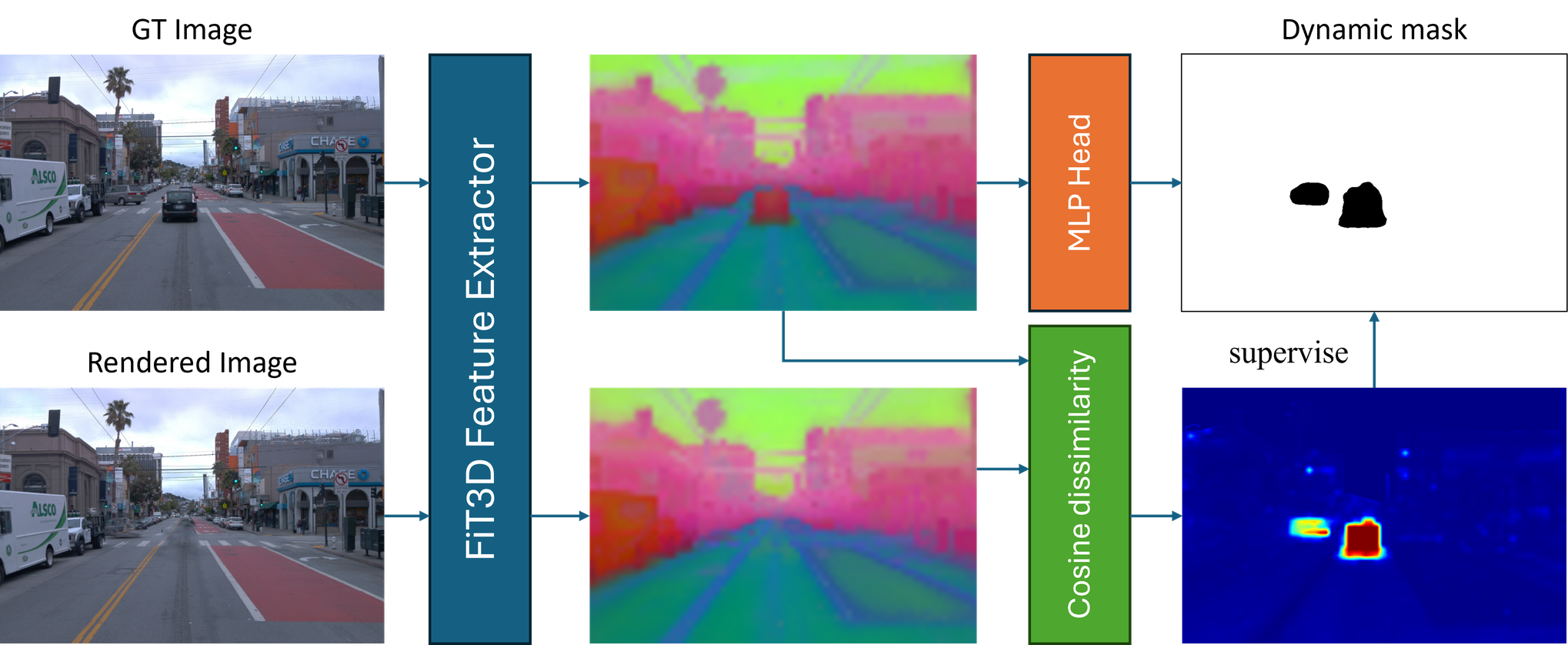}
    \caption{Segmentation Mask Extraction of \ourmethod. We utilize the differences between the rendered image and ground truth to train a dynamic mask decoder.}
    \label{fig:mask}
\end{figure*}
\begin{figure*}
    \centering
    \includegraphics[width=\linewidth]{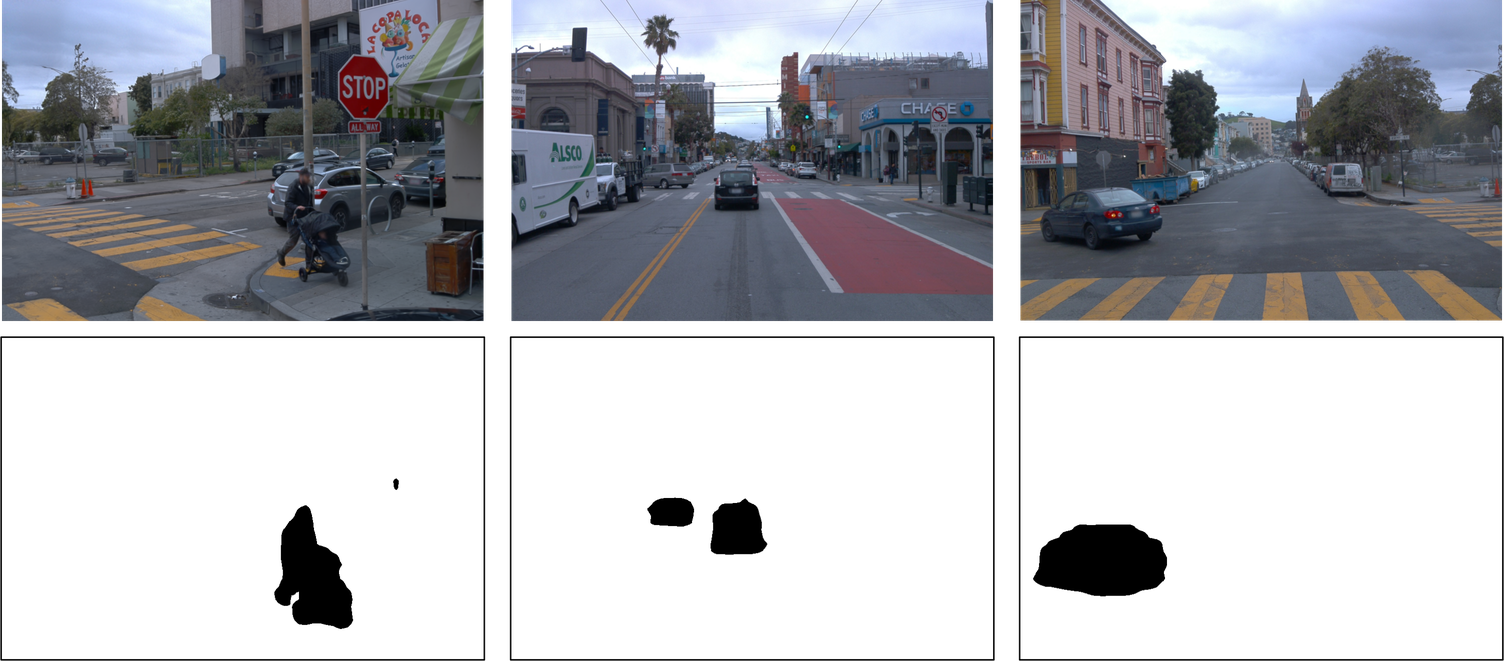}
    \caption{Extracted motion masks using FiT3D features}
    \label{fig:extracted-motion-masks}
\end{figure*}

\vspace{5pt}
\noindent \textbf{Feature Extractor.}
DINOv2 \cite{dinov2_2023_arxiv} is widely used as a foundation model for feature extraction and have demonstrated potential in previous novel view synthesis works, such as Wild-Gaussians \cite{wildgaussians_2024_nips}.
However,  we found through experiments that the features extracted from DINOv2 are usually noisy, especially on the road and in the sky, as shown in Fig. \ref{fig:dino-vs-fit3d}. The DINOv2 features cannot produce accurate motion masks sometimes. On the other hand, FiT3D \cite{fit3d_2024_eccv} fine-tunes DINOv2 with gaussian splatting to improve the 3D awareness of the extracted features, which is a perfect match for our setting in the driving world. Therefore, we turn to FiT3D as the feature extractor, producing more clean and robust features, to measure the similarity of two images. 

\vspace{5pt}
\noindent \textbf{Motion Mask Extractor.}

By introducing the learnable decoder, we are not limited to the viewpoints where GT images are available. Instead,  given a rendered image, we can first extract the FiT3D features, and then use our decoder to extract the motion mask, without the requirement for ground truth images.

During training, the joint optimization of image rendering and mask prediction will benefit from each other by using the obtained mask $M$ to mask out the dynamic regions. The rendering loss is as follows:
\begin{equation}
    \mathcal{L}_{masked-render} = M \odot\| \hat{I} - I \|
\end{equation}

As we mask out the dynamic regions, the reconstruction at the regions will not be supervised. As a result, the difference between the rendered images  and the ground truth images will be become more significant, which benefits the extraction of the desired motion masks.

We provide a few samples in Fig. \ref{fig:extracted-motion-masks}. It can be observed that our model can handle the dynamic objects well, even for far-away pedestrians.

\vspace{5pt}
\noindent \textbf{Temporal Geometric Constraints.}
Due to the sparsity nature of views in driving scenarios, it tends to overfit to the training views when optimizing gaussian splatting. Single-view image loss often suffers from texture-less area in far distance. As a result, relying on photometric consistency is not reliable. Instead, we propose to enhance the geometric consistency by aggregating temporal information. 

Based on the assumption that depth of static regions remains consistent across time from varying views, we designate a cross-view temporal spatial consistency module. For a static pixel $(u_r, v_r)$ in the reference frame, with depth value of $d_r$, we can project it to the nearest neighboring view, which has the largest amount of overlap. Given the camera intrinsics $K$ and extrinsics $T_r, T_n$, we can obtain the corresponding pixel location in the neighboring view:
\begin{equation}
[u_n, v_n, 1]^T = K T_n T_r^{-1} \left(d_r \cdot K^{-1}  [u_r, v_r, 1]^T \right)
\end{equation}

Again, we can query the depth value $d_n$ at the position $(u_n, v_n)$. When we project it back to the 3D space, the position should be consistent with the one obtained from back-projecting $(u_r, v_r, d_r)$ to the reference frame.
\begin{equation}
[u_{nr}, v_{nr}, 1]^T = K T_r  T_n^{-1} \left(d_n \cdot  K^{-1}[u_n, v_n, 1]^T \right)
\end{equation}

We apply geometric loss to optimize the Gaussians to produce cross-view consistent depth as follows:
\begin{equation}
\label{eq:geo-reg2}
    \mathcal{L}_{uv} = \| (u_r, v_r) - (u_{nr}, v_{nr}) \|_2
\end{equation}

\begin{figure*}
    \centering
    \includegraphics[width=\linewidth]{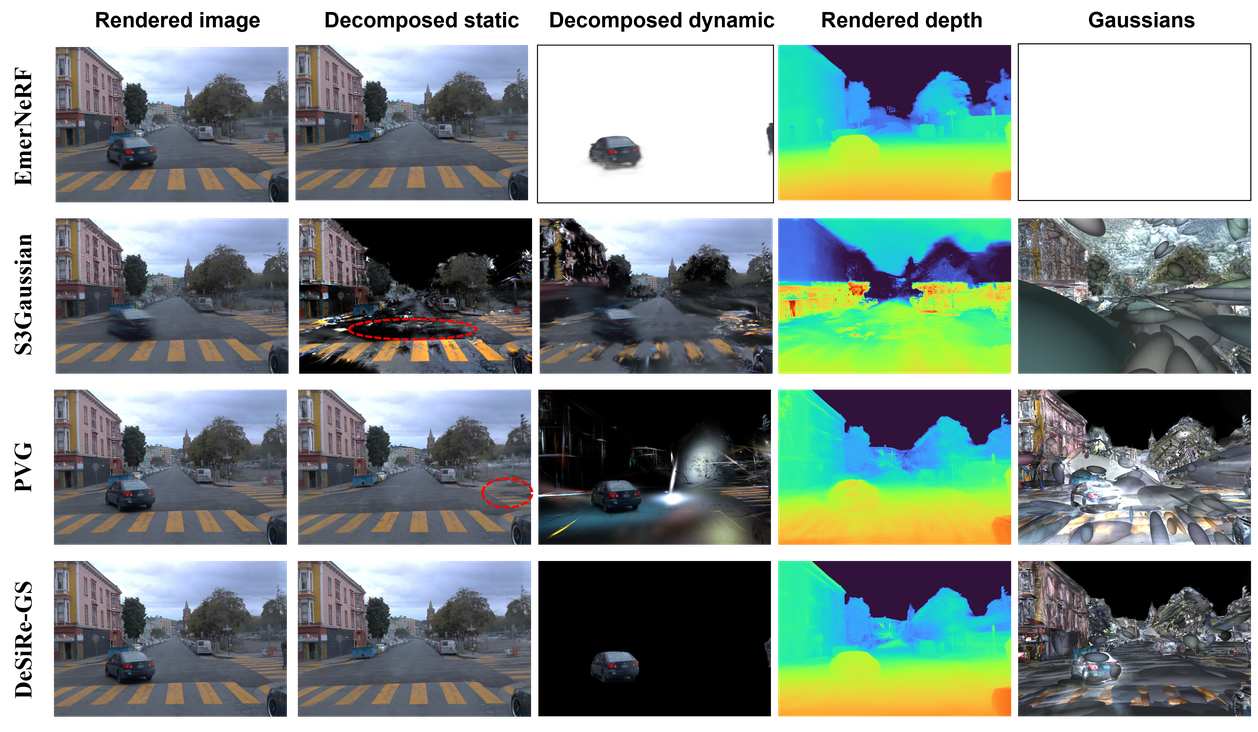}
    \caption{Qualitative Comparison.}
    \label{fig:depth-comp}
\end{figure*}
\section{Baselines}

\begin{itemize}
    \item[$\bullet$] \textbf{StreetSurf}~\cite{streetsurf_2023_arxiv} is an implicit neural rendering method for both geometry and appearance reconstruction in street views. The whole scene is divided in to close-range, distant-view and sky parts according to the distance of objects. A cuboid close-range hash grid and a hyper-cuboid distant-view model are employed to tackle long and narrow observation space in most street scenes, showcasing good performance in unbounded scenes captured by long camera trajectories.
    \item[$\bullet$] \textbf{NSG}~\cite{nsg_2021_cvpr} enables efficient rendering of novel arrangements and views by encoding object transformations and radiance within a learnable scene graph representation. It contains a background node approximating all static parts, several dynamic nodes representing rigidly moving individuals, and edges representing transformations. NSG\cite{nsg_2021_cvpr} also combines implicitly encoded scenes with a jointly learned latent representation to describe objects in a single implicit function. 
    \item[$\bullet$] \textbf{SUDS}~\cite{suds_2023_cvpr} is a NeRF-based method for dynamic large urban scene reconstruction. It proposed using 2D optical flow to model scene dynamics, avoiding additional bounding box annotations. SUDS develops a three-branch hash table representation for 4D scene representation, enabling a variety of downstream tasks. 
    \item[$\bullet$] \textbf{StreetGS}~\cite{streetgs_2024_eccv} models dynamic driving scenes using 3D Gaussian splatting. It represents the components in the scene separately, with a background model for static part and an object model for foreground moving objects. To capture dynamic features, the position and rotation of gaussians are defined in an object local coordinate system, which relies on bounding boxes predicted by an off-the-shelf model.
    \item[$\bullet$] \textbf{HUGS}~\cite{hugs_2024_cvpr} is a 3DGS-based method addressing the problem of urban scene reconstruction and understanding. It assumes that the scene is composed of static rigions and moving vehicles with rigid motions, using a unicycle model to model vehicles' states. HUGS also extends original 3DGS to model additional modalities, including optical flow and semantic information, achieving good performance in both scene reconstruction and semantic reconstruction. Bounding boxes are also required in this process.
    \item[$\bullet$] \textbf{EmerNeRF}~\cite{emernerf_2023_arxiv} is a NeRF-based method for constructing 4D neural scene representations in urban driving scenes. It decomposes dynamic scenes into a static field and a dynamic field, both parameterized by hash grids. Then an emergent scene flow field is introduced to represent explicit correspondences between moving objects and aggregate temporally-displaced features. Remarkably, EmerNeRF finishes these tasks all through self-supervision.
    \item[$\bullet$] \textbf{S3Gaussian}~\cite{s3g_2024_arxiv} is a self-supervised approach that decomposes static and dynamic 3D gaussians in driving scenes. It aggregates 4D gaussian representations in a spatial-temporal field network with a multi-resolution hexplane encoder, where the dynamic objects are visible only within spatial-temporal plane while static objects within spatial-only plane. Then S3Gaussian utilizes a multi-head decoder to capture the deformation of 3D Gaussians in a canonical space for decomposition. 
    \item[$\bullet$] \textbf{Omnire}~\cite{omnire_2024_arxiv} successfully models urban dynamic scenes using Gaussian Scene Graphs, with different types of nodes tackling sky, background, rigidly moving objects and non-rigidly moving objects. It introduces rigid nodes for vehicles, where the Gaussians will not change over time, and non-rigid nodes for human-ralated dynamics, where local deformations will be taken into consideration. OmniRe additionally employs a Skinned Multi-Person Linear (SMPL) model to parameterize human body model, showcasing good results in reconstructing in-the-wild humans. Notably, Omnire also requires accurate instance bounding boxes for dynamic modeling.
    \item[$\bullet$] \textbf{PVG}~\cite{pvg_2023_arxiv} is a self-supervised gaussian splatting approach that reconstructs dynamic urban scenes and isolates dynamic parts from static background. Refer to Sec. \ref{sec:related_work} for more details about PVG.
\end{itemize}

In the approaches mentioned above, StreetSurf\cite{streetsurf_2023_arxiv}, Mars\cite{mars_2023_caai},SUDS\cite{suds_2023_cvpr} and EmerNeRF\cite{emernerf_2023_arxiv} are based upon NeRF, while others are based upon 3DGS. Notably, among the 3DGS-based approaches, HUGS\cite{hugs_2024_cvpr}, StreetGS\cite{streetgs_2024_eccv} and OmniRe\cite{omnire_2024_arxiv} all rely on instance-level bounding boxes for moving objects, which are sometimes difficult to obtain. PVG\cite{pvg_2023_arxiv} and S3Gaussian\cite{s3g_2024_arxiv} are most closely related to our work, both of which are self-supervised Gaussian Splatting method without reliance on extra annotations.

\begin{table*}[t]
    \centering
    \resizebox{\textwidth}{!}{%
    \begin{tabular}{lcccccccccc}
        \toprule
       \multirow{3}{*}{\textbf{Method}}  & \multicolumn{6}{c}{\textbf{Dynamic-32 Split}} & \multicolumn{4}{c}{\textbf{Static-32 Split}} \\
    \cmidrule(lr){2-7} \cmidrule(lr){8-11} 
         & \multicolumn{3}{c}{{Image reconstruction}} & \multicolumn{3}{c}{{Novel view synthesis}} &  \multicolumn{2}{c}{{Image reconstruction}} & \multicolumn{2}{c}{{Novel view synthesis}}  \\
         & PSNR $\uparrow$ & DPSNR $\uparrow$ & L1$\downarrow$
         & PSNR $\uparrow$ & DPSNR $\uparrow$ & L1$\downarrow$
         & PSNR $\uparrow$ &  L1$\downarrow$
        & PSNR $\uparrow$ &  L1$\downarrow$ \\
        \midrule
        3DGS~\cite{3dgs_2023_TOG}  & 28.47 & 23.26 & - & 25.14 & 20.48 & - & 29.42 & - & 26.82& -\\
        Mars~\cite{mars_2023_caai}  & 28.24 & 23.37 & - & 26.61 & 22.21 & - & 28.31& - & 27.63& - \\
        EmerNeRF~\cite{emernerf_2023_arxiv}  & 28.16 & 24.32 & 3.12& 25.14 & 23.49 & 4.33&  30.00& 2.84& 28.89& 3.89\\
        S3Gaussian~\cite{s3g_2024_arxiv}  & 31.35 & 26.02 & 5.31& 27.44 & 22.92 & 6.18& 30.73& 5.84& 27.05& 6.53\\  
        PVG \cite{pvg_2023_arxiv} & 33.14 & 31.79 & 3.33& 29.77& 27.19& 4.84& 32.84& 3.75& 29.12& 5.07\\
        \midrule
        Ours  & 34.56 & 32.63 & 2.96& 30.45 & 28.66& 4.17& 34.57& 2.89& 31.78& 3.93\\
        \bottomrule
    \end{tabular}
    }
        \caption{Comparison of methods on the Waymo NOTR Dataset from EmerNeRF.}
    \label{tab:notr_results}
\end{table*}

\begin{table*}[t]
    \centering
    \resizebox{\textwidth}{!}{%
    \begin{tabular}{lccccccccc}
        \toprule
        \textbf{Segment Name} & seg104554 & seg125050 & seg169514 & seg584622 & seg776165 & seg138251 & seg448767 & seg965324 \\
        \midrule
        \textbf{Scene Index} & 23 & 114 & 327 & 621 & 703 & 172 & 552 & 788 \\
        \bottomrule
    \end{tabular}
    }
    \caption{Segment Names and Scene IDs of 8 scenes used in OmniRe\cite{omnire_2024_arxiv}.}
    \label{tab:omnire_segment}
\end{table*}
\begin{table}[thpb]
    \centering
    \resizebox{\linewidth}{!}{%
    \begin{tabular}{lcccc}
        \toprule
        \textbf{Segment Name} & seg102319 & seg103913 & seg109636 & seg117188 \\
        \midrule
        \textbf{Scene Index} & 17 & 22 & 50 & 81 \\
        \bottomrule
    \end{tabular}
    }
    \caption{Segment Names and Scene IDs of 4 scenes used in PVG\cite{pvg_2023_arxiv}.}
    \label{tab:pvg_segment}
\end{table}

\section{Data}

We conduct our experiments on the Waymo Open Dataset \cite{waymo_2020_cvpr} and the KITTI Dataset \cite{kitti_2012_cvpr}, both consisting of real-world autonomous driving scenarios.

\subsection{Waymo Open Dataset}

\vspace{5pt}
\noindent \textbf{NOTR from EmerNeRF.} NOTR is a subset consisting of diverse and balance sequences derived from Waymo Open Dataset introduced by \cite{emernerf_2023_arxiv}. It includes 120 distinct driving sequences, categorized into 32 static, 32 dynamic, and 56 diverse scenes covering various challenging driving conditions. Following \cite{s3g_2024_arxiv}, we incorporate the 32 dynamic scenes and 32 static scenes from NOTR into our testing set. Refer to EmerNeRF\cite{emernerf_2023_arxiv} for NOTR dataset details.

\vspace{5pt}
\noindent \textbf{OmniRe subset.} OmniRe\cite{omnire_2024_arxiv} selects eight highly complex dynamic driving sequences from Waymo Open Dataset, each including dynamic classes such as vehicles and pedestrains. The Segment IDs of selected scenes are shown in Tab.\ref{tab:omnire_segment}. 

\vspace{5pt}
\noindent \textbf{PVG subset.} PVG \cite{pvg_2023_arxiv} provides four sequences randomly selected from Waymo Open Dataset, which are also included in our experiments. The Segment IDs are shown in Tab.\ref{tab:pvg_segment}.

For the sequences in the Waymo dataset, we follow the same setup as \cite{emernerf_2023_arxiv}. Camera images are captured from three frontal cameras—FRONT LEFT, FRONT, and FRONT RIGHT—and then resized to a resolution of \(640 \times 960\). Only the first return of the LiDAR point cloud data is considered. We select the first 50 frames from each dataset for our experiments, and then scale the time range to [0,1].

\subsection{KITTI Dataset}

For KITTI Dataset, we only test DeSiRe-GS on the subset provided by PVG\cite{pvg_2023_arxiv}. 
Different from the Waymo dataset, we only use the left and right cameras for evaluation on KITTI dataset. The resolution of images from each camera is \(375 \times 1242\). Similar to Waymo Dataset preprocessing, we randomly choose 50 frames from the whole sequence from each KITTI dataset and rescale time duration to [0,1] with a frame interval of 0.02 seconds.
\subsection{Data Source}
For Tab. \ref{tab:results}, the results of the baselines are taken from PVG \cite{pvg_2023_arxiv}, since we are using the same dataset and evaluated on the devices. The results of Tab. \ref{tab:omnire} are sourced from OmniRe \cite{omnire_2024_arxiv}.

\section{Additional Results}

\subsection{Additional quantitative results}
We conducted experiments on the NOTR dataset, and the results are listed in Tab. \ref{tab:notr_results}. Following PVG \cite{pvg_2023_arxiv}, we scaled the camera pose and point clouds during pre-processing. For a fair comparison of depth errors with other methods, we re-map the depth back to the original scale and calculate the depth L1 error. The results in Tab. \ref{tab:ablations} for ablation studies are without the rescaling.



\subsection{Analysis}

We compare the rendered depth of various methods in Fig. \ref{fig:depth-comp}. 
S3Gaussian \cite{s3g_2024_arxiv} fails to predict accurate depth map, because they use only LiDAR point clouds for initialization, where there are no points at the upper part. 
Other than the LiDAR points, PVG \cite{pvg_2023_arxiv} and \ourmethod randomly sample points, enabling us to render much better depth map.

GS-based methods, such as PVG \cite{pvg_2023_arxiv} and S3Gaussian \cite{s3g_2024_arxiv} generally outperform NeRF-based methods like EmerNeRF \cite{emernerf_2023_arxiv} in terms of image rendering. However, the explicit GS methods tend to overfit to the images, thereby performing poorly on depth rendering. With the proposed cross-view consistency, our model can successfully solve the over-fitting problem, achieving satisfactory rendering quality both in image and depth.

\end{document}